\def\year{2020}\relax
\documentclass[letterpaper]{article} % DO NOT CHANGE THIS
\usepackage{aaai20}  % DO NOT CHANGE THIS
\usepackage{times}  % DO NOT CHANGE THIS
\usepackage{helvet} % DO NOT CHANGE THIS
\usepackage{courier}  % DO NOT CHANGE THIS
\usepackage[hyphens]{url}  % DO NOT CHANGE THIS
\usepackage{graphicx} % DO NOT CHANGE THIS
\usepackage{graphicx}  % DO NOT CHANGE THIS
\usepackage{bm,color,paralist,url}
\usepackage{subfig}
\usepackage{enumitem}

\usepackage{algorithm,algpseudocode}

\usepackage{epstopdf}

\usepackage{amsmath}
\usepackage{amsfonts}

\newcommand{\ie}{{i.e.}}

\newcommand{\hide}[1]{}

\newcommand{\norm}[1]{\left\lVert#1\right\rVert}

\newcommand{\Lp}{\left(}
\newcommand{\Rp}{\right)}

\newcommand*{\Scale}[2][4]{\scalebox{#1}{$#2$}}

\newtheorem{theorem}{Theorem}
\newtheorem{proposition}{Proposition}
\newtheorem{lemma}{Lemma}
\newtheorem{corollary}{Corollary}
\newtheorem{problem}{Problem}
\newtheorem{assumption}{Assumption}
\newtheorem{proof}{Proof}

\title{Stable Prediction with Model Misspecification and Agnostic Distribution Shift \thanks{All authors of this paper are corresponding authors.}}
\author{Kun Kuang\textsuperscript{\rm 1,2}\thanks{Most of this work was done while Kun Kuang was a Ph.D. student in Tsinghua University}, Ruoxuan Xiong\textsuperscript{\rm 3}, Peng Cui\textsuperscript{\rm 2}, Susan Athey\textsuperscript{\rm 3}, Bo Li\textsuperscript{\rm 2}\\
\textsuperscript{\rm 1}Zhejiang University\\
\textsuperscript{\rm 2}Tsinghua University\\
\textsuperscript{\rm 3}Stanford University\\ %If you have multiple authors and multiple affiliations
kunkuang@zju.edu.cn, rxiong@stanford.edu, cuip@tsinghua.edu.cn\\
athey@stanford.edu, libo@sem.tsinghua.edu.cn % email address must be in roman text type, not monospace or sans serif
}
 
\begin{document}

\maketitle

\begin{abstract}
For many machine learning algorithms, two main assumptions are required to guarantee performance. One is that the test data are drawn from the same distribution as the training data, and the other is that the model is correctly specified.
In real applications, however, we often have little prior knowledge on the test data and on the underlying true model.
Under model misspecification, agnostic distribution shift between training and test data leads to inaccuracy of parameter estimation and instability of prediction across unknown test data.
To address these problems, we propose a novel Decorrelated Weighting Regression (DWR) algorithm which jointly optimizes a variable decorrelation regularizer and a weighted regression model.
The variable decorrelation regularizer estimates a weight for each sample such that variables are decorrelated on the weighted training data. Then, these weights are used in the weighted regression to improve the accuracy of estimation on the effect of each variable, thus help to improve the stability of prediction across unknown test data.
Extensive experiments clearly demonstrate that our DWR algorithm can significantly improve the accuracy of parameter estimation and stability of prediction with model misspecification and agnostic distribution shift.

\end{abstract}

\section{Introduction}
Predicting unknown outcomes based on a model estimated on a training data set is a common machine learning problem.
Many machine learning algorithms have been proposed and shown to be very successful for prediction when the test data have the same distribution as the training data or the model specification is correct. %if model correct, don't need training = test distribution
In real applications, however, we rarely know the underlying true model for prediction, and we cannot guarantee the unknown test data will have the same distribution as the training data. For example, different geographies, schools, or hospitals may draw from different demographics, and the correlation structure among demographics may also vary (e.g. one ethnic group may be more or less disadvantaged in different geographies). If the model is misspecified, it may exploit subtle statistical relationships among features present in the training data to improve prediction, resulting in inaccuracy of parameter estimation and instability of prediction across test data sets with different distributions.

To correct the distribution shift between training and test data, many methods have been proposed in domain adaption \cite{bickel2009discriminative,ben2010theory} and transfer learning \cite{pan2010survey}. The motivation of these methods is to adjust the distribution of training data to mimic the distribution of test data, so that a predictive algorithm trained on training data can minimize the predictive error on test data. These methods achieve good performance in practice, however, they require the test data as prior knowledge. Hence, they cannot be applied to address the agnostic distribution shift problem.

Recently, some algorithms have been proposed to address the agnostic distribution shift from unknown test data, including domain generalization \cite{muandet2013domain}, causal transfer learning \cite{rojas2018invariant} and invariant causal prediction \cite{peters2016causal} etc.
Their motivation is to explore the invariant structure between predictors and the response variable across multiple training datasets for prediction.
But they cannot well handle the case of distribution shifts that are not observed in the training data. Moreover, they do not consider the interaction of distribution shift and model misspecification. Recently, some papers \cite{kuang2018stable,shen2018causally} were proposed to address stable prediction problem using methods drawn from the literature on causal inference, achieving improved performance. But they did not consider the model misspecification and their algorithms were restricted to the predictive setting with binary predictors and binary response variable.

In this paper, we focus on the problem of stable prediction with model misspecification and agnostic distribution shift, where we assume that all features (predictors) $\mathbf{X}$ fall into one of two categories: one category includes the stable features $\mathbf{S}$, which have causal effects on outcome $Y$ that are invariant across environments (e.g., across training and test sets).
%In a general prediction problem, all features $\mathbf{X}$ fall into one of two categories: one category is the stable features $\mathbf{S}$, whose causal effects on outcome $Y$ are non-zero and invariant across datasets.
%For example, ears, noses and whiskers are stable features to recognize a cat in an image.
%The other category is the unstable features $\mathbf{V}$, which have no causal effects on outcome, but would be correlated with either stable features, the outcome, or both since data bias.
The other category includes the unstable features $\mathbf{V}$, which have no causal effects on outcome, but may be correlated with either stable features, the outcome, or both. The correlation may be different in different environments.
%For example, the background (\ie, grass) is unstable feature, but might be highly correlated with cat in some image data.
Under the assumption that all stable features are observed, model misspecification would be induced by some omitted nonlinear or interaction terms of stable features (\ie, $s_1\cdot s_2$ or $e^{s_1\cdot s_2}$).
Since different environments (e.g., training and test sets) have different covariate distributions, the parameters estimated from different environments may be quite different even when we use the same parametric model. This variation in parameters arises because the parameters on included features capture two components: first, the partial effect of the included features on the expected value of outcome, and second, a function that depends on the correlation between included and omitted features, as well the distribution of outcomes conditional on omitted features.  We consider the the problem of making predictions when that second component of estimated parameters is unstable across environments.  In that case, we prefer to find an estimator that eliminates the second component, even though including it improves prediction for test sets that are similar to the training data.  We look for an algorithm that is effective when the analyst does not know which feature is stable feature and which is not.
% Then, sampling bias might bring spurious correlation between the missing nonlinear or interaction term  and unstable features, resulting in inaccuracy on parameter estimation, hence instability on prediction.

One way to address the problem of stable prediction in such a setting is to isolate the impact of each individual feature.  One method commonly used in the causal literature is covariate balancing \cite{athey2018approximate,kuang2017treatment,kuang2017estimating}, which essentially estimates the impact of the target feature by reweighting the data so that the distribution of covariates is equalized across different values of the target feature.  This literature usually focuses on the case where there is a single, pre-specified feature of interest and other features are considered to be ``confounders''.
In this paper, we consider the case where there are potentially many stable features, and propose a novel Decorrelated Weighting Regression (DWR) algorithm for stable prediction with model misspecification and agnostic distribution shift by jointly optimizing a variable decorrelation regularizer and a weighted regression model.
Specifically, the variable decorrelation regularizer constructs sample weights to reduce correlation among covariates and allows the weighted regression to approximately isolate the effect of each variable.
The weighted regression model with those sample weights might perform worse than standard methods when predicting in the test data with similar distribution to the training, but it will do better across unknown test data with distribution shift from the training.
%By simultaneously optimize variables decorrelation and the weighted regression, we can learn a more precise and robust predictive algorithm for stable prediction across unknown environments.
Using both empirical experiments and theoretical analysis, we show that our algorithm outperforms alternatives in parameter estimation and stability in prediction across unknown test data.

This paper has three main contributions: 1) we investigate the problem of stable prediction with model misspecification and agnostic distribution shift. The problem setting is more general and practical than prior work. 2) We propose a novel DWR algorithm to jointly optimize variable decorrelation and weighted regression to address the stable prediction problem. 3) We conduct extensive experiments in both synthetic and real-world datasets to demonstrate the advantages of our algorithm on stable prediction problem.

\section{Problem and Our Algorithm}
In this section, we first give the formulation of stable prediction problem, then introduce the details of our algorithm.

\subsection{Stable Prediction Problem}

\hide{
 with following definitions:
\begin{eqnarray}
\label{metrics:acc} \!\!\!\!\!\! \Scale[0.9]{Average\_Error} \!\!\!\!\!\! &=& \!\!\!\!\!\! \Scale[0.9]{\frac{1}{|\mathcal{E}|}\sum_{e \in \mathcal{E}}Error(D^e)},\\
\label{metrics:stb} \!\!\!\!\!\!  \Scale[0.9]{Stability\_Error} \!\!\!\!\!\! &=& \!\!\!\!\!\! \Scale[0.9]{\sqrt{\frac{1}{|\mathcal{E}|-1}\sum_{e \in \mathcal{E}}\left(Error(D^e)-Average\_Error\right)^{2}}},
\end{eqnarray}
where $|\mathcal{E}|$ refers to the number of environments, and $Error(D^e)$ represents the predictive error on dataset $D^e$.
Actually, $Average\_Error$ and $Stability\_Error$ refer to the mean and variance of the predictive error over all possible environment $e\in \mathcal{E}$.
}

Let $\mathcal{X}$ denote the space of observed features and $\mathcal{Y}$ denote the outcome space.
We define an \textbf{environment} to be a joint distribution $P_{XY}$ on $\mathcal{X} \times \mathcal{Y}$, and let $\mathcal{E}$ denote the set of all environments.
In each environment $e\in \mathcal{E}$, we have dataset $D^{e} = (\mathbf{X}^{e}, Y^{e})$, where $\mathbf{X}^e \in \mathcal{X}$ are predictor variables and $Y^e \in \mathcal{Y}$ is a response variable.  The joint distribution of features and outcomes on $(\mathbf{X},Y)$ can change across environments: $P^{e}_{XY} \neq P^{e'}_{XY}$ for $e,e'\in\mathcal{E}$.

In this paper, our goal is to learn a predictive model for stable prediction with model misspecification and agnostic distribution shift. To measure its performance on stable prediction problem, we adopt the $Average\_Error$ and $Stability\_Error$ in \cite{kuang2018stable} as:
\begin{eqnarray}
\label{def:average} \Scale[0.9]{Average\_Error} \!\!\!\!\!\! &=& \!\!\!\!\!\! \Scale[0.9]{\frac{1}{|\mathcal{E}|}\sum\limits_{e \in \mathcal{E}}RMSE(D^e)},\\
\label{def:stability} \Scale[0.9]{Stability\_Error} \!\!\!\!\!\! &=& \!\!\!\!\!\! \Scale[0.9]{\sqrt{\frac{1}{|\mathcal{E}|-1}\sum\limits_{e \in \mathcal{E}}\left(RMSE(D^e)-Average\_Error\right)^{2}},}
\end{eqnarray}
where $|\mathcal{E}|$ refers to the number of test environments, and $RMSE(D^e)$ represents the Root Mean Square Error of a predictive model on dataset $D^e$. Actually, $Average\_Error$ and $Stability\_Error$ refer to the mean and variance of predictive error over all possible environments $e \in \mathcal{E}$.

Then, the stable prediction problem \cite{kuang2018stable} is defined as:
\begin{problem}[Stable Prediction]
\textbf{Given} one training environment $e\in \mathcal{E}$ with dataset $D^{e}=(\mathbf{X}^{e},Y^{e})$, the task is to \textbf{learn} a predictive model to predict across unknown environment $\mathcal{E}$ with not only small $Average\_Error$ but also small $Stability\_Error$.
\end{problem}

Letting $\mathbf{X} = \{\mathbf{S}, \mathbf{V}\}$, we define $\mathbf{S}$ as stable features, and $\mathbf{V}$ as unstable features with Assumption \ref{asmp:stable}:
\begin{assumption}
\label{asmp:stable}
There exists a stable function f(s) such that for all environment $e\in \mathcal{E}$, $\mathbb{E}(Y^e|\mathbf{S}^e=s, \mathbf{V}^e = v) = \mathbb{E}(Y^e|\mathbf{S}^e=s) = f(s)$.
\end{assumption}

Assumption \ref{asmp:stable} can be guaranteed by $Y\perp\mathbf{V}|\mathbf{S}$.
Thus, we can address the stable prediction problem by developing a predictive model that learns the stable function $f(\mathbf{S})$.
But we have NO prior knowledge on which features are stable and which are unstable.
\begin{assumption}
\label{asmp:allfeatures}
All stable features $\mathbf{S}$ are observed.
\end{assumption}

Under Assumption \ref{asmp:allfeatures}, model misspecification will be induced when estimating an outcome function if the model omits some nonlinear transformations and interaction terms of the stable features.
%By supposing that the misspecified model is linear, then the true stable function $f(\mathbf{S})$ will be a non-linear function with following generative model for the outcome in environment $e$:
%Model misspecification will be induced by missing some nonlinear transformations and interaction terms of the stable features.
Suppose that the true stable function $f(\mathbf{S})$ and $Y$ in environment $e$ is given by:
\begin{eqnarray}
Y^e = f(\mathbf{S}^e)+\mathbf{V}^e\beta_V+\epsilon^e = \mathbf{S}^e\beta_S + g(\mathbf{S}^e) +\mathbf{V}^e\beta_V+ \varepsilon^e.
\end{eqnarray}
where $\beta_V = 0$ and $\varepsilon^e \perp \mathbf{X}^e$. We assume that the analyst misspecifies the model by omitting $g(\mathbf{S}^e)$ and uses a linear model for prediction.

Under Assumption \ref{asmp:stable}, the distribution shift across environments is mainly induced by the variation in the joint distribution over $(\mathbf{V}^e,\mathbf{S}^e)$. Simple linear regression may estimate nonzero effects of unstable features $\mathbf{V}^e$ when $\mathbf{V}^e$ is correlated with the omitted variables $g(\mathbf{S}^e)$. For OLS, we have
\begin{eqnarray}
\nonumber \Scale[0.9]{\hat{\beta}_{V_{OLS}} } &=&  \Scale[0.9]{\beta_V+(\frac{1}{n}\sum\limits_{i=1}^n \mathbf{V}_i^T\mathbf{V}_i)^{-1}(\frac{1}{n}\sum\limits_{i=1}^n \mathbf{V}_i^Tg(\mathbf{S}_i))}\\
\label{eq:V_OLS} &+&\Scale[0.9]{(\frac{1}{n}\sum\limits_{i=1}^n \mathbf{V}_i^T\mathbf{V}_i)^{-1}(\frac{1}{n}\sum\limits_{i=1}^n \mathbf{V}_i^T\mathbf{S}_i)(\beta_{S} - \hat{\beta}_{S_{OLS}})},\\
\nonumber \Scale[0.9]{\hat{\beta}_{S_{OLS}}}&=& \Scale[0.9]{\beta_S+(\frac{1}{n}\sum\limits_{i=1}^n \mathbf{S}_i^T\mathbf{S}_i)^{-1}(\frac{1}{n}\sum\limits_{i=1}^n \mathbf{S}_i^Tg(\mathbf{S}_i))}\\
\label{eq:S_OLS} &+&\Scale[0.9]{(\frac{1}{n}\sum\limits_{i=1}^n \mathbf{S}_i^T\mathbf{S}_i)^{-1}(\frac{1}{n}\sum\limits_{i=1}^n \mathbf{S}_i^T\mathbf{V}_i)(\beta_{V} - \hat{\beta}_{V_{OLS}})},
\end{eqnarray}
where $n$ is sample size, $\frac{1}{n}\sum_{i=1}^n \mathbf{V}_i^Tg(\mathbf{S}_i) = \mathbb{E}(\mathbf{V}^T g(\mathbf{S}))+o_p(1)$ and $\frac{1}{n}\sum_{i=1}^n \mathbf{V}_i^T\mathbf{S}_i =  \mathbb{E} (\mathbf{V}^T \mathbf{S})+o_p(1)$. To simplify notation, we remove the environment variable $e$ from $\mathbf{X}^e$, $\mathbf{S}^e$, $\mathbf{V}^e$, $\varepsilon^e$.

If $\mathbb{E} (\mathbf{V}^T g(\mathbf{S})) \neq 0$ or $\mathbb{E} (\mathbf{V}^T \mathbf{S}) \neq 0$ in Eq. (\ref{eq:V_OLS}), $\hat{\beta}_{V_{OLS}} $ will be biased, resulting in the biased estimation on $\mathbf{S}$ in Eq. (\ref{eq:S_OLS}). And its prediction will be very unstable since the correlation between $\mathbf{V}$ and $g(\mathbf{S})$ (or $\mathbf{S}$) might vary across testing environments. Hence, to increase the stability of prediction, we need to precisely estimate the parameters of $\hat{\beta}_{V_{OLS}} $ by removing the correlation between $\mathbf{V}$ and $g(\mathbf{S})$ (or $\mathbf{S}$) on training data, that is let $\mathbb{E} (\mathbf{V}^T g(\mathbf{S})) = 0$ and $\mathbb{E} (\mathbf{V}^T \mathbf{S}) = 0$.

\hide{
For OLS, the non-linear part $g(\mathbf{S}^e)$ can be regarded as omitted variable, and by assuming that the mean value of , we have:
\begin{eqnarray}
\label{eq:V_OLS}\hat{\beta}_{V_{OLS}} \!\!\!\!\!\! &=& \!\!\!\!\!\! \beta_V+\frac{Cov(\mathbf{V},g(\mathbf{S}))}{Var(\mathbf{V})}+\frac{Cov(\mathbf{V},\mathbf{S})}{Var(\mathbf{V})}(\beta_{S} - \hat{\beta}_{S_{OLS}}),\\
\label{eq:S_OLS}\hat{\beta}_{S_{OLS}} \!\!\!\!\!\! &=& \!\!\!\!\!\! \beta_S+\frac{Cov(\mathbf{S},g(\mathbf{S}))}{Var(\mathbf{S})}+\frac{Cov(\mathbf{S},\mathbf{V})}{Var(\mathbf{S})}(\beta_{V} - \hat{\beta}_{V_{OLS}}),
\end{eqnarray}
where we remove the environment variable $e$ from $\mathbf{X}^e$, $\mathbf{S}^e$, $\mathbf{V}^e$, $\varepsilon^e$ for notations simplification.
}

\hide{
If the correlation between $\mathbf{V}$ and $g(\mathbf{S})$ (or $\mathbf{S}$) varies across environments, instability arises.
On the other hand, if $\mathbb{E} (\mathbf{V}^T g(\mathbf{S})) = 0$ and $\mathbb{E} (\mathbf{V}^T \mathbf{S}) = 0$ in a given environment, the estimated coefficients will be stable across environments, and $\hat{\beta}_{V_{OLS}}$ will be approximately zero.\footnote{We can normalize the mean value of $\mathbf{X} = \{\mathbf{S},\mathbf{V}\}$ to be $zero$. We can also include a constant variable (the intercept) in regression.}
}

\hide{
To simplify notations, we remove the environment variable $e$ from $\mathbf{X}^e$, $\mathbf{S}^e$, $\mathbf{V}^e$, $\varepsilon^e$, and $Y^e$ when there is no confusion from the context.
}

%\noindent

\textbf{Notations.} In our paper, $n$ refers to the sample size, and $p$ is the dimensions of variables. For any vector $\textbf{v} \in \mathbb{R}^{p\times 1}$, let $\|\textbf{v}\|_2^2 = \sum_{i=1}^{p}v_i^{2}$, and $\|\textbf{v}\|_1 = \sum_{i=1}^{p}|v_i|$. For any matrix $\mathbf{X}\in \mathbb{R}^{n\times p}$, we let $\mathbf{X}_{i,}$ and $\mathbf{X}_{,j}$ represent the $i^{th}$ sample and the $j^{th}$ variable in $\mathbf{X}$, respectively.

\subsection{Variable Decorrelation}

In this subsection, we introduce our variable decorrelation regularizer to reduce the correlation between $\mathbf{V}$ and  $\mathbf{S}$ (or $g(\mathbf{S})$) in the training environment.

\begin{proposition}
\label{pro:mutually_independent}
If $\mathbf{X}$ are mutually independent with mean 0, then $\mathbb{E} (\mathbf{V}^T g(\mathbf{S})) = 0$ and $\mathbb{E} (\mathbf{V}^T \mathbf{S}) = 0$.
\end{proposition}

Proposition \ref{pro:mutually_independent} together with Eq. (\ref{eq:V_OLS}) and Eq. (\ref{eq:S_OLS}) imply that if the covariates are mutually independent, we can unbiasedly estimate parameter $\beta_{V}$ even $g(\mathbf{S})$ is omitted.  This motivates our regularizer.

From \cite{Bisgaard2006}, we know variables $\mathbf{X}_{,j}$ and $\mathbf{X}_{,k}$ are independent if $\mathbb{E}[\mathbf{X}_{,j}^a\mathbf{X}_{,k}^b] = \mathbb{E}[\mathbf{X}_{,j}^a]\mathbb{E}[\mathbf{X}_{,k}^b]$ for all $a,b\in \mathbb{N}$.\footnote{In empirical applications, we can
discretize $\mathbf{X}_{,j}$ and $\mathbf{X}_{,k}$ to satisfy the sufficient condition in \cite{Bisgaard2006}.}
Inspired by the weighting methods in the causal literature \cite{athey2018approximate,fong2018covariate,kuang2017estimating}, we propose to make $\mathbf{X}_{,j}$ and $\mathbf{X}_{,k}$ become independent by reweighting samples with weights $W$, which can be learnt with the following objective function:
\begin{eqnarray}
\label{eq:independent_W}
\Scale[1.0]{\min\limits_{W}\sum\limits_{a = 1}^{\infty}\sum\limits_{b = 1}^{\infty}\| \mathbb{E}[\mathbf { X } _ {,j }^{a^T}\mathbf{\Sigma}_W\mathbf { X } _ {  , k }^{b}] -  \mathbb{E}[\mathbf { X } _ { , j }^{a^T}W] \mathbb{E}[\mathbf { X } _ { , k }^{b^T}W] \|_2^2},
\end{eqnarray}
where $W\in \mathbb{R}^{n\times 1}$ are sample weights, $\sum_{i=1}^n W_i = n$ and $\mathbf{\Sigma}_W = diag(W_{1},\cdots,W_{n})$ is the corresponding diagonal matrix.
%To remove those spurious correlation induced by $\mathbb{E} (\mathbf{V}^T g(\mathbf{S})) \neq 0$ and $\mathbb{E} (\mathbf{V}^T \mathbf{S}) \neq 0$, one need to make all variables in $\mathbf{X}$ become mutually independent.
In practice, however, it will not be feasible to attain the objective that all the moments of variables in the objective function from Eq. (\ref{eq:independent_W}) are equal to zero.
Fortunately, from Eq. (\ref{eq:V_OLS}) and Eq. (\ref{eq:S_OLS}) we know that reducing correlation among the first moments of the variables can help to improve the precision of parameter estimation and the stability of predictive models, and in practice the analyst can include high-order moments, for example, polynomial functions of covariates to further improve stability.

In this paper, we focus on variables' first moment and propose to de-correlate all the predictors by sampling reweighting in the training environment.
Specifically, we propose a \emph{variable decorrelation} regularizer for learning that sample weight $W$ as follows:
\begin{eqnarray}
\label{eq:our_weighting} \Scale[0.9]{W^b  = \arg \min\limits_ { W } \sum_ { j = 1 } ^ { p }  \norm{ \mathbb{E}[\mathbf { X } _ {,j }^T\mathbf{\Sigma}_W\mathbf { X } _ {  , - j }] -  \mathbb{E}[\mathbf { X } _ { , j }^TW] \mathbb{E}[\mathbf { X } _ { , - j }^TW] }_2^2}
\end{eqnarray}
\hide{
\begin{eqnarray}
\nonumber \Scale[0.9]{W} \!\!\!\! &=& \!\!\!\! \Scale[0.9]{\arg \min\limits_ { W } \sum\limits_ { j = 1 } ^ { p }  \norm{ \mathbb{E}[\mathbf { X } _ {,j }^TW\mathbf { X } _ {  , - j }] -  \mathbb{E}[\mathbf { X } _ { , j }^TW] \mathbb{E}[\mathbf { X } _ { , - j }^TW] }_2^2}\\
\label{eq:our_weighting} \!\!\!\! &=& \!\!\!\! \Scale[0.9]{\arg \min\limits_ { W } \sum\limits_ { j = 1 } ^ { p }  \norm{ \mathbf { X } _ {,j }^TW\mathbf { X } _ {  , - j }/n -  \mathbf { X } _ { , j }^TW/n \cdot \mathbf { X } _ { , - j }^TW/n }_2^2 }
\end{eqnarray}
}
where $\mathbf { X } _ {  , - j } = \mathbf { X } \backslash \{ \mathbf { X }_{,j} \}$ means all the remaining variables by removing the $j^{th}$ variable in $\mathbf{X}$.\footnote{We obtain $\mathbf{X}_{,-j}$ in experiment by setting the value of $j^{th}$ variable in $\mathbf{X}$ as $zero$.}
The summand represents the loss due to correlation between variable $\mathbf{X}_{,j}$ and all other variables $\mathbf{X}_{,-j}$.
Note that, only first moment is considered in Eq. (\ref{eq:our_weighting}), but it is sufficient for variables decorrelation. And higher-order moments can be easily incorporated.

The following theoretical results (proved in the supplementary material) show that our variable decorrelation regularizer can make the variables in $\mathbf{X}$ become mutually uncorrelated by sample reweighting, hence reduce the correlation among covariates in the training environment and improve the accuracy on parameter estimation.

With $\sum_{i=1}^n W_i = n$, we can denote the loss in Eq. (\ref{eq:our_weighting}) as:
\begin{eqnarray}
\Scale[0.9]{\mathcal{L}_{B} =  \sum_ { j = 1 } ^ { p }  \norm{ \mathbf { X } _ {,j }^T\mathbf{\Sigma}_W\mathbf { X } _ {  , - j }/n -  \mathbf { X } _ { , j }^TW/n \cdot \mathbf { X } _ { , - j }^TW/n }_2^2 }.
\end{eqnarray}

\begin{lemma}
\label{lemma:ideal_weights}
If the number of covariates $p$ is fixed, then there exists a sample weight $ W\succeq 0$ such that
\begin{equation}
\label{eq:theorem_balancing}
\lim\limits_{n\rightarrow \infty} \mathcal{L}_B = 0
\end{equation}
with probability $1$. In particular, a solution $W$ to Eq. (\ref{eq:theorem_balancing}) is $W_{i}^{\star} = \frac{\Pi_{j=1}^{p} \hat{f}(\mathbf{X}_{i,j})}{\hat{f}(\mathbf{X}_{i,1},\cdots,\mathbf{X}_{i,p})}$, where $\hat{f}(x_{\cdot,j}) $ and $\hat{f}(x_{\cdot,1},\cdots,x_{\cdot,p})$ are the Kernel density estimators.\footnote{In detail, $\hat{f}(x_{i,j}) = \frac{1}{nh_j} \sum_{i = 1}^n  k \left( \frac{\mathbf{X}_{i,j} - x_{i,j}}{h_j} \right)$, where $k(u)$ is a kernel function and $h_j$ is the bandwidth parameter for covariate $\mathbf{X}_j$; and $\hat{f}(x_{i}) = \frac{1}{n|H|} \sum_{i = 1}^n K \left( H^{-1} (\mathbf{X}_i - x_i) \right)$, where $K(u)$ is a multivariate kernel function, $H = diag(h_1, \cdots, h_p)$ and  $|H| = h_1\cdots h_p$.}
\hide{, that is,  $\hat{f}(x_{i,j}) = \frac{1}{nh_j} \sum_{i = 1}^n  k \left( \frac{\mathbf{X}_{i,j} - x_{i,j}}{h_j} \right)$, where $k(u)$ is a kernel function and $h$ is the bandwidth parameter; and $\hat{f}(x_{i}) = \frac{1}{n|H|} \sum_{i = 1}^n K \left( H^{-1} (\mathbf{X}_i - x_i) \right)$, where $K(u)$ is a multivariate kernel function, $H = diag(h_1, \cdots, h_p)$ and  $|H| = h_1\cdots h_p$.}
\end{lemma}
\begin{proof}
See Appendix.
\end{proof}

But the solution $W$ that satisfies Eq. (\ref{eq:theorem_balancing}) in Lemma \ref{lemma:ideal_weights} is not unique.
To address this problem, we propose to simultaneously minimize the variance of $W$ and restrict the sum of $W$ in our regularizer as follows:
\begin{eqnarray}
\label{eq:loss_W}
\hat{W} = \arg \min\limits_ { W \in \mathcal{C}} \mathcal{L}_B  + \frac{ \lambda_3 }{n}\Scale[1.0]{\sum_{i=1}^n W_{i}^2} +  \lambda_4 (\frac{1}{n} \Scale[1.0]{\sum_{i=1}^{n}W_{i}- 1})^2,
\end{eqnarray}
where $\mathcal{C} = \{W: |W_{ij}| \leq c\}$ for some constant $c$.
%The restriction $\left(\frac{1}{n} \sum_{i=1}^{n}W_{i}- 1\right)^2$ avoids all the sample weights to be $zero$.
%$\frac{1}{n} \sum_{i=1}^{n}W_{i} = 1$ matches the total new sample weights to the original total equal weights.

Then, we have following theorem on our variable decorrelation regularizer in Eq. (\ref{eq:loss_W}).
\begin{theorem}
\label{theo:unique_solution}
The solution $\hat{W}$ defined in Eq. (\ref{eq:loss_W}) is unique if $\lambda_3 n\gg p^2 + \lambda_4$, $p^2 \gg \max(\lambda_3, \lambda_4)$ and $|\mathbf{X}_{i,j}| \leq c$ for some constant $c$.
\end{theorem}
\begin{proof}
See Appendix.
\end{proof}

With Lemma \ref{lemma:ideal_weights} and Theorem \ref{theo:unique_solution}, we can derive the following property of the $\hat{W}$ given by Eq. (\ref{eq:loss_W}).

\hide{
\begin{corollary}
\label{coro:W_deterministic_function}
The sample weight $\hat{W}$ learned by our global variable balancing regularizer in Eq. \ref{eq:loss_W} can be represented a deterministic function of $\mathbf{X}$ under Theorem \ref{theo:unique_solution}. Formally, $\hat{W} = g(\mathbf{X})=g(\mathbf{S},\mathbf{V})$.
\end{corollary}
}

\textbf{Property 1.} \textit{When $p$ is fixed, $n\rightarrow \infty$, $\lambda_3 n\gg p^2 + \lambda_4$, and $p^2 \gg \max(\lambda_3, \lambda_4)$, the variables in $\mathbf{X}$ become uncorrelated by sample reweighting with $\hat{W}$.  Hence, correlation between $\mathbf{V}$ and $\mathbf{S}$ in the training environment will be removed.}
\hide{
\begin{proposition}
\nonumber
\label{theo:uncorrelated}
When $p$ is fixed and $n\rightarrow \infty$, the variables in $\mathbf{X}$ become uncorrelated after sample reweighting by $\hat{W}$.
Hence, the correlation between $\mathbf{V}$ and $\mathbf{S}$ will be removed, say $Cov(\mathbf{S}, \mathbf{V}) = 0$, while the correlation between $\mathbf{V}$ and $g(\mathbf{S})$ will also be reduced to some degree.
\end{proposition}
}

Extensive empirical experiments demonstrate that the correlation between $\mathbf{V}$ and $g(\mathbf{S})$ will also be reduced by our regularizer.  In summary, the proposed variable decorrelation regularizer in Eq. (\ref{eq:loss_W}) can learn a unique optimal sample weights $\hat{W}$ that can de-correlate the variables $\mathbf{X}$, and thus improve the accuracy in parameters estimation and stability in prediction.

\subsection{Decorrelated Weighting Regression}

With the learned sample weights $\hat{W}$ from variable decorrelation regularizer in Eq. (\ref{eq:loss_W}), one can run weighted least square (WLS) to estimate the regression coefficient $\beta$ as:
\begin{eqnarray}
\label{eq:WLS}
\hat{\beta}_{WLS} = \arg \min\limits_{\beta} \sum\limits_{i=1}^{n} \hat{W}_{i}\cdot (Y_i-\mathbf{X}_{i,}\beta)^2.
\end{eqnarray}
The $\hat{\beta}_{WLS}$ is expected to have less bias than $\hat{\beta}_{OLS}$ under Property 1, since sample reweighted by $\hat{W}$ de-correlates variables in $\mathbf{X}$.

By combining the objective functions of the variable decorrelation regularizer in Eq. (\ref{eq:loss_W}) and the weighted regression in Eq. (\ref{eq:WLS}), we propose a Decorrelated Weighted Regression (DWR) algorithm to jointly optimize sample weights $W$ and regression coefficient $\beta$ as follows:
\begin{eqnarray}
\label{eq:our}
 \!\!\!\!\!\!\!\! && \Scale[1.0]{\min\limits_{W,\beta} \sum_{i=1}^{n} W_{i}\cdot (Y_i-\mathbf{X}_{i,}\beta)^2}\\
\nonumber \!\!\!\!\!\!\!\! &s.t& \Scale[1.0]{\sum _ { j = 1 } ^ { p } \norm{ \mathbf { X } _ {,j }^T\mathbf{\Sigma}_W\mathbf { X } _ {  , - j }/n -  \mathbf { X } _ { , j }^TW/n \cdot \mathbf { X } _ { , - j }^TW/n }_2^2 < \lambda_2}\\
\nonumber \!\!\!\!\!\!\!\!&&\Scale[1.0]{|\beta|_1<\lambda_1, \ \ \frac{1}{n} \sum_{i=1}^{n}W_{i}^2<\lambda_3},\\
\nonumber \!\!\!\!\!\!\!\!&&(\frac{1}{n} \sum_{i=1}^{n} W_{i} -1)^2<\lambda_4, \ \  W\succeq 0,
\end{eqnarray}
where $n$ denotes the sample size, $p$ refers to the dimension of variables $\mathbf{X}$. $\mathbf{X}_{i,}$ and $\mathbf{X}_{,j}$ represent the $i^{th}$ sample and the $j^{th}$ variable in $\mathbf{X}$, respectively.
The term $W\succeq 0$ constrains each sample weight to be non-negative. With term $\frac{1}{n}\sum_{i=1}^{n}W_{i}^2<\lambda_3$, we reduce the variation of the sample weights. The term $(\frac{1}{n} \sum_{i=1}^{n} W_{i} -1)^2<\lambda_4$ avoids all sample weights to be $zero$.

\begin{algorithm}[tbp]
\caption{{Decorrelated Weighted Regression algorithm}}
\label{alg:dwr}
\begin{algorithmic}[1]
\Require
Observed features $\mathbf{X}$ and outcome variable $Y$.
\Ensure
Updated parameters $W$, $\beta$.
\State Initialize parameters  $W^{(0)}$ and $\beta^{(0)}$,
\State Calculate loss function with parameters $(W^{(0)},\beta^{(0)})$,
\State Initialize the iteration variable $t\leftarrow 0$,

\Repeat
\State $t\leftarrow t+1$,
\State Update $W^{(t)}$ with gradient descent by fixing $\beta$,
\State Update $\beta^{(t)}$ with gradient descent by fixing $W$,
\State Calculate loss function with parameters $(W^{(t)},\beta^{(t)})$,
\Until{Loss function converges or max iteration is reached}.\\
\Return $W$, $\beta$.
\end{algorithmic}
\end{algorithm}

\section{Optimization and Analysis}

\subsection{Optimization}
To optimize our DWR algorithm in Eq. (\ref{eq:our}), we propose an iterative method.
Firstly, we initialize sample weights
%$W_{i} = 1/n$
$W_{i} = 1$
for each sample $i$ and regression coefficient $\beta = [0,0,\cdots,0]^T$.
Once the initial values are given, in each iteration, we first update $W$ by fixing $\beta$, then update $\beta$ by fixing $W$ until the objective function in Eq. (\ref{eq:our}) converges.
The whole algorithm is summarized in Algorithm \ref{alg:dwr}.

\subsection{Complexity Analysis}
In our DWR algorithm, the main time cost is to calculate the value of loss function and update parameters $W$ and $\beta$ in each iteration.
The complexity of calculating the loss function is $O(np^2)$, where $n$ is the sample size and $p$ refers to the dimension of observed variables.
The complexity of updating parameter $W$ is also $O(np^2)$.
The complexity of updating parameter $\beta$ is $O(np)$.

In total, the complexity of each iteration in Algorithm \ref{alg:dwr} is $O(np^2)$.

\section{Experiments}
In this section, we check the performance of our algorithm with experiments on both synthetic and real-world datasets.
\subsection{Baselines}
We use following four methods as the baselines.
\begin{itemize}%[leftmargin=0.7cm]
\item \par \noindent Ordinary Least Square (OLS):
$$\min \|Y-\mathbf{X}\beta\|_2^2.$$
\item \par \noindent Lasso \cite{tibshirani1996regression}:
$$\min \|Y-\mathbf{X}\beta\|_2^2+\lambda_1\|\beta\|_1.$$
\item \par \noindent Ridge Regression \cite{hoerl1970ridge}:
$$\min \|Y-\mathbf{X}\beta\|_2^2+\lambda_1\|\beta\|_2.$$
\item \par \noindent Independently Interpretable Lasso (IILasso) \cite{takada2017independently}
$$\min \|Y-\mathbf{X}\beta\|_2^2+\lambda_1\|\beta\|_1+\lambda_2|\beta|^T\mathbf{R}|\beta|,$$ where $\mathbf{R} \in \mathcal{R}^{p\times p}$ with each element $\mathbf{R}_{jk}=|r_{jk}|/(1-|r_{jk}|)$, and $r_{jk} = \frac{1}{n}|\mathbf{X}_{,j}^T\mathbf{X}_{,k}|$.
\end{itemize}
To avoid the degeneration of above baselines, we set their hype-parameters $\lambda_1 \neq 0$ and $\lambda_2\neq 0$.

\hide{
In experiments, we compare our algorithm with following baselines, including OLS, LASSO \cite{tibshirani1996regression}, Ridge Regression \cite{hoerl1970ridge} and Independently Interpretable Lasso (IILasso) \cite{takada2017independently}.
}

\subsection{Evaluation Metrics}
To evaluate the performance of stable prediction, we use $RMSE$, $\beta\_Error$, $Average\_Error$ and $Stability\_Error$ as evaluation metrics. Their definitions of $RMSE$ and $\beta\_Error$ are listed as follows:
\begin{eqnarray}
\Scale[1.0]{RMSE = \sqrt{\frac{1}{n}\sum_{k=1}^{n}(Y_k-\hat{Y}_k)}},
\end{eqnarray}
where $n$ is sample size, $\hat{Y}_k$ and $Y_k$ refer to the predicted and true outcome for sample $k$.
\begin{eqnarray}
\Scale[1.0]{\beta\_Error = \|\beta-\hat{\beta}\|_1},
\end{eqnarray}
where $\hat{\beta}$ and $\beta$ represent the estimated and true regression coefficients.
\hide{
\begin{eqnarray}
 \Scale[0.9]{Average\_Error} \!\!\!\!\!\! &=& \!\!\!\!\!\! \Scale[0.9]{\frac{1}{|\mathcal{E}|}\sum_{e \in \mathcal{E}}RMSE(D^e)},\\
 \Scale[0.9]{Stability\_Error} \!\!\!\!\!\! &=& \!\!\!\!\!\! \Scale[0.9]{\sqrt{\frac{1}{|\mathcal{E}|-1}\sum_{e \in \mathcal{E}}\left(RMSE(D^e)-Average\_Error\right)^{2}},}
\end{eqnarray}
where $|\mathcal{E}|$ refers to the number of test environments, and $RMSE(D^e)$ represents the RMSE value on dataset $D^e$ from environment $e$.
}
\subsection{Experiments on Synthetic Data}

\subsubsection{Dataset}

\begin{figure}[tb]
\centering
\subfloat[$\mathbf{S}\perp \mathbf{V}$ \label{fig:s0v}]{
  \includegraphics[width=1.0in]{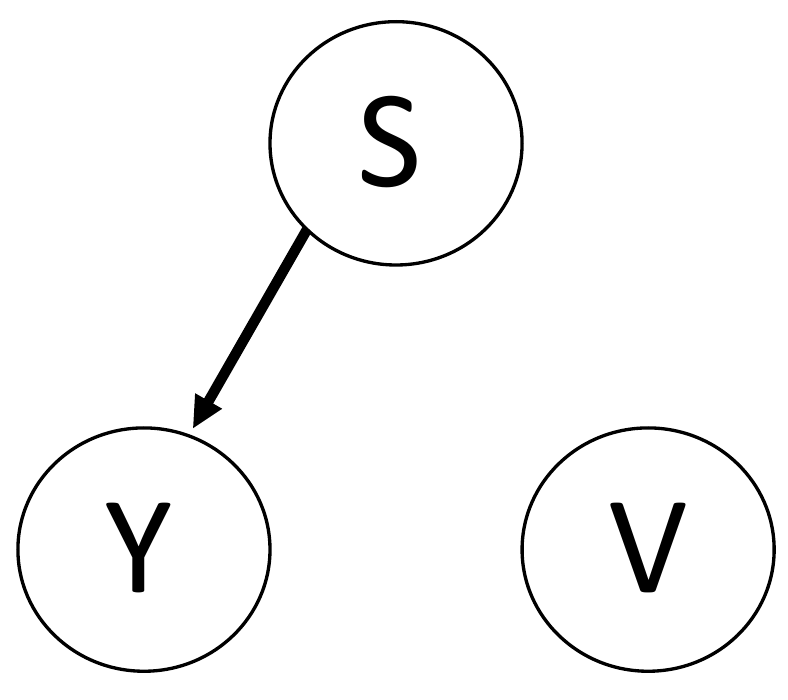}
}
\subfloat[$\mathbf{S}\rightarrow \mathbf{V}$ \label{fig:s2v}]{
  \includegraphics[width=1.0in]{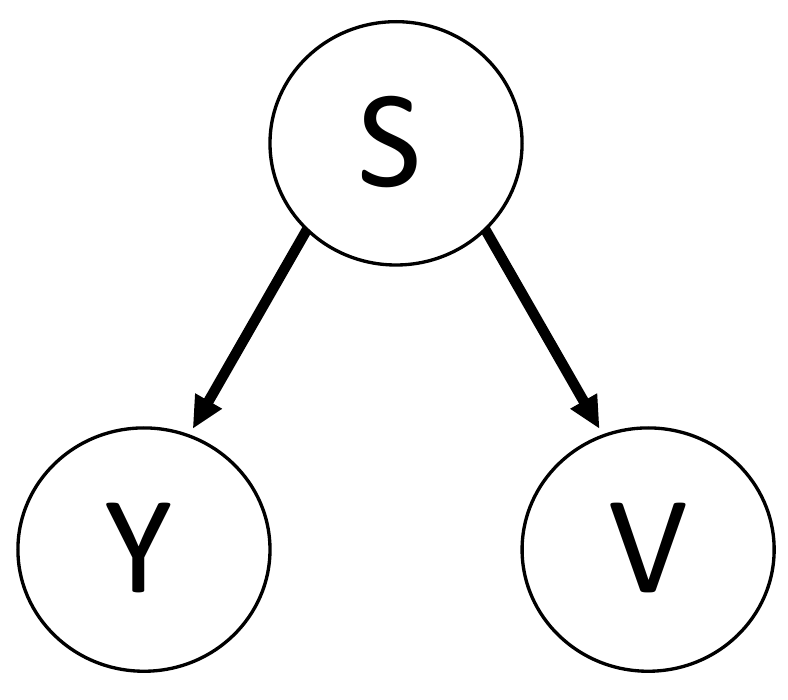}
}
\subfloat[$\mathbf{V}\rightarrow \mathbf{S}$ \label{fig:v2s}]{
  \includegraphics[width=1.0in]{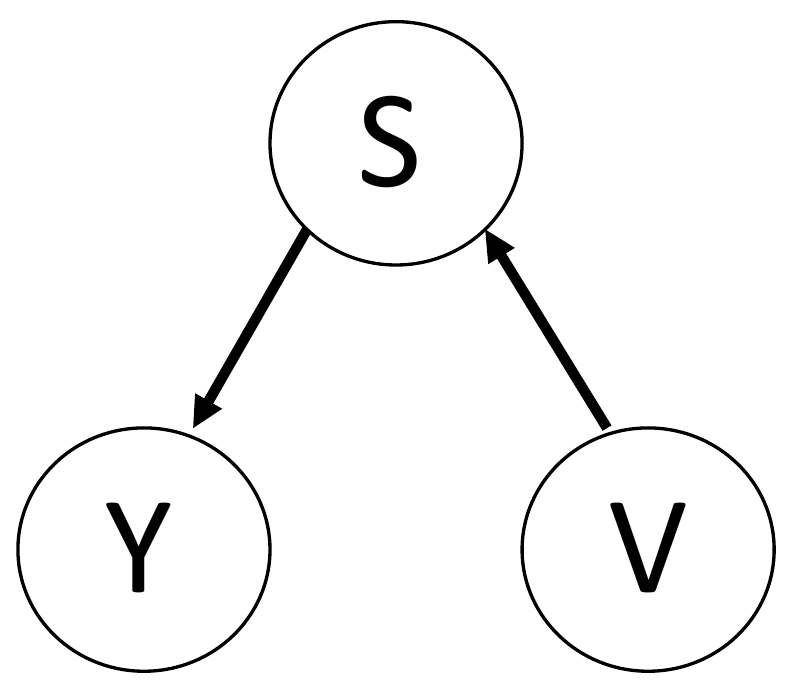}
}
\caption{Three diagrams for stable features $\mathbf{S}$, unstable features $\mathbf{V}$, and response variable $Y$.}
\label{fig:graph}
\end{figure}

Under Assumption \ref{asmp:stable}, there are three kinds of relationships between $\mathbf{X} = \{\mathbf{S}, \mathbf{V}\}$ and $Y$ as shown in Fig. \ref{fig:graph}, including $\mathbf{S}\perp \mathbf{V}$, $\mathbf{S}\rightarrow \mathbf{V}$, and $\mathbf{V}\rightarrow \mathbf{S}$.
We consider settings motivated by each of the three cases as follows:

\noindent \textbf{$\mathbf{S}\perp \mathbf{V}$:} In this setting, $\mathbf{S}$ and $\mathbf{V}$ are independent, but $\mathbf{S}$ could be dependent with each other.
Hence, we generate $\mathbf{X} =\{\mathbf{S}_{,1}, \cdots, \mathbf{S}_{,p_s}, \mathbf{V}_{,1}, \cdots, \mathbf{V}_{,p_v}\}$ with independent Gaussian distributions with the help of auxiliary variables $\mathbf{Z}$ as following:
\begin{eqnarray}
\label{eq:V_independent}\Scale[1.0]{\mathbf{Z}_{,1}, \cdots, \mathbf{Z}_{,p}\   \overset{iid}{\sim} \mathcal{N}(0,1)},  \Scale[1.0]{\mathbf{V}_{,1}, \cdots, \mathbf{V}_{,p_v}\   \overset{iid}{\sim} \mathcal{N}(0,1)} \\
\label{eq:S_dependent}\mathbf{S}_{,i} = 0.8*\mathbf{Z}_{,i} + 0.2*\mathbf{Z}_{,i+1}, \,\,  i = 1, 2, \cdots, p_s,
\end{eqnarray}
where the number of stable variables $p_s = 0.5*p$ and the number of unstable variables $p_v = 0.5*p$. $\mathbf{S}_{,j}$ represents the $j^{th}$ variable in $\mathbf{S}$.

\noindent \textbf{$\mathbf{S}\rightarrow \mathbf{V}$:} In this setting, the stable features $\mathbf{S}$ are the causes of unstable features $\mathbf{V}$.
We first generate dependent stable features $\mathbf{S}$ with Eq. (\ref{eq:S_dependent}). Then, we generate unstable features $\mathbf{V}$ based on $\mathbf{S}$:
$\Scale[1.0]{\mathbf{V}_{\cdot,j} = 0.8*\mathbf{S}_{\cdot,j}+0.2*\mathbf{S}_{\cdot,j+1}+\mathcal{N}(0,1)}$,
where we let $j+1 = mod(j+1,p_s)$. The function $mod(a,b)$ returns the modulus after division of $a$ by $b$.

\noindent \textbf{$\mathbf{V}\rightarrow \mathbf{S}$:} In this setting, unstable features $\mathbf{V}$ are the causes of stable features $\mathbf{S}$. We first generate the unstable features $\mathbf{V}$ with Eq. (\ref{eq:V_independent}).
Then, we generate the stable features $\mathbf{S}$ based on $\mathbf{V}$:
$\Scale[1.0]{\mathbf{S}_{\cdot,j} = 0.2*\mathbf{V}_{\cdot,j}+0.8*\mathbf{V}_{\cdot,j+1}+\mathcal{N}(0,1)}$,
where we let $j+1 = mod(j+1,p_v)$.

To test the performance with different forms of missing nonlinear and interaction terms, we generate the outcome $Y$ from a polynomial nonlinear function ($Y_{poly}$) and an exponential one ($Y_{exp}$):
\begin{eqnarray}
\label{eq:Y_simulation_poly} \!\!\!\!\!\! Y_{poly} \!\!\!\!\!\! &=& \!\!\!\!\!\! f(\mathbf{S}) + \varepsilon  = [\mathbf{S},\mathbf{V}]\cdot [\beta_s, \beta_v]^T + \mathbf{S}_{\cdot,1}\mathbf{S}_{\cdot,2}\mathbf{S}_{\cdot,3} + \varepsilon,\\
\label{eq:Y_simulation_exp} \!\!\!\!\!\! Y_{exp} \!\!\!\!\!\! &=& \!\!\!\!\!\! f(\mathbf{S}) + \varepsilon  = [\mathbf{S},\mathbf{V}]\cdot [\beta_s, \beta_v]^T + e^{\mathbf{S}_{\cdot,1}\mathbf{S}_{\cdot,2}\mathbf{S}_{\cdot,3}} + \varepsilon.
\end{eqnarray}
where $\beta_s = \{\frac{1}{3},-\frac{2}{3},1,-\frac{1}{3},\frac{2}{3},-1,\cdots\}$, $\beta_v = \vec 0$, and $\varepsilon =  \mathcal{N}(0,0.3)$.
%where $\beta_s = \{\underbrace{1,-2,3,-1,2,-3,\cdots}_{p_s}\}$, and $\beta_v = \{\underbrace{0,0,\codts,0}_{p_v}\}$.

\begin{figure}[t]
\centering
\subfloat[On raw data]{
  \includegraphics[width=1.4in]{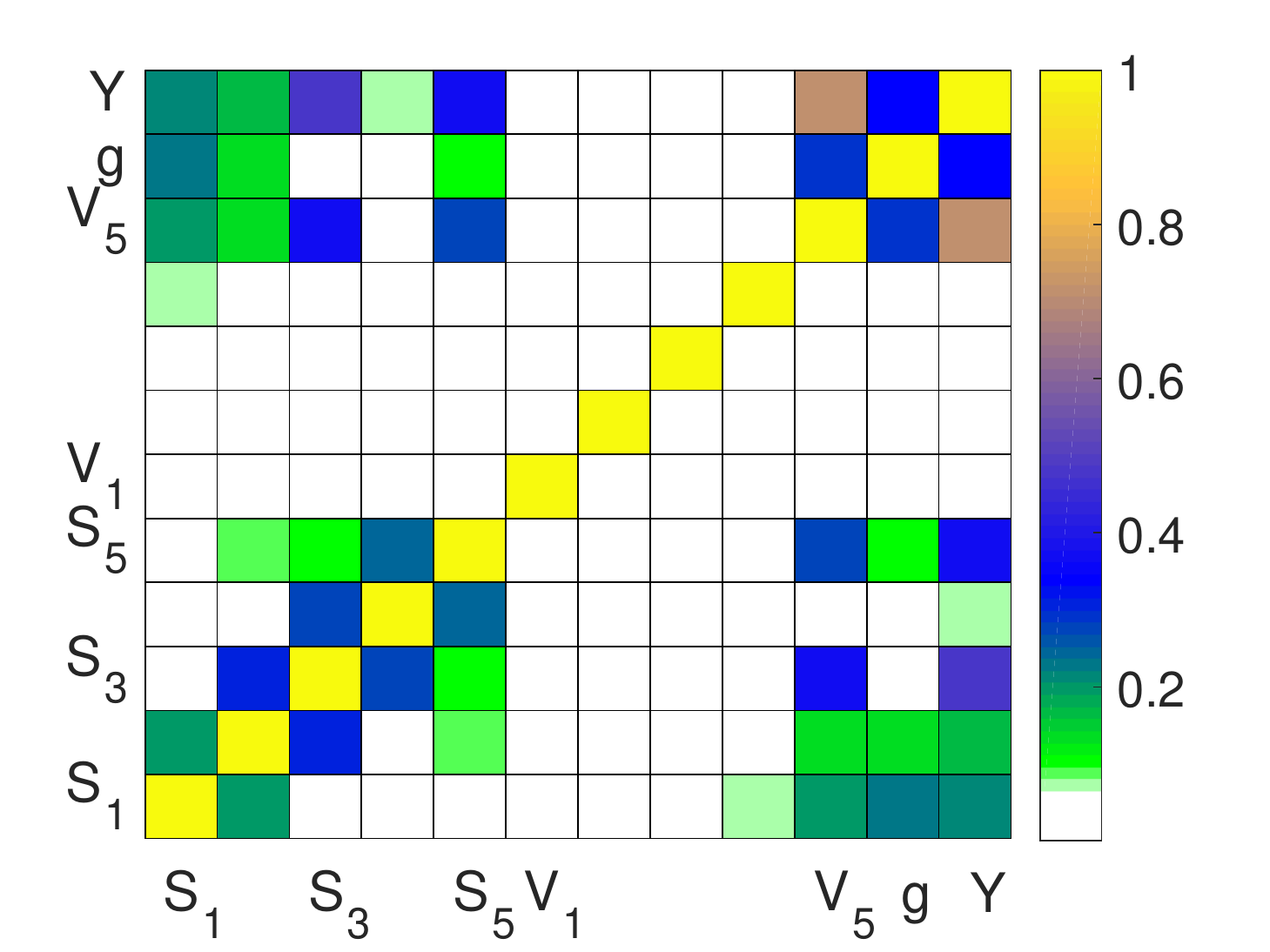}
}
\subfloat[On the weighted data]{
  \includegraphics[width=1.4in]{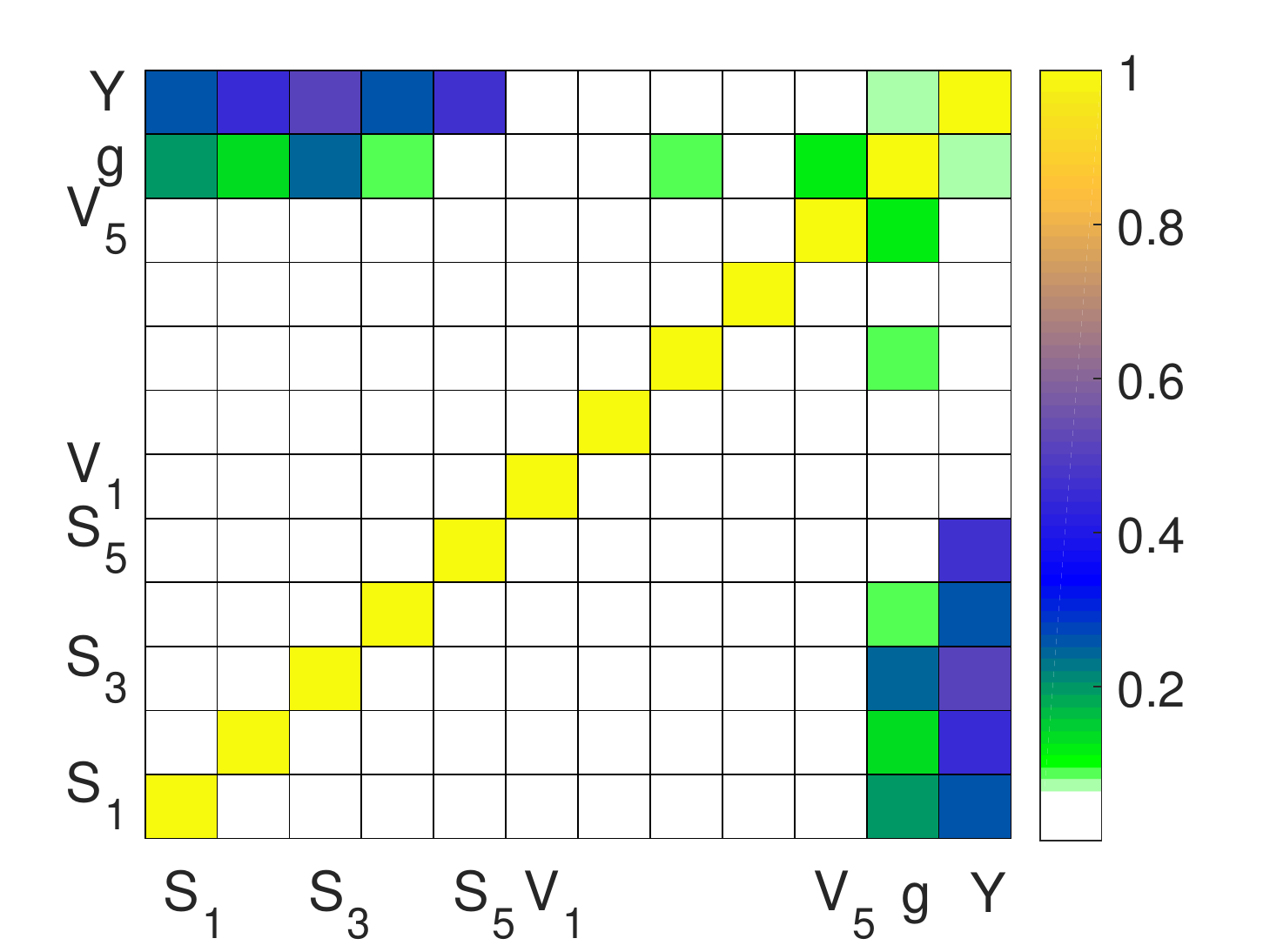}
}
\caption{Pearson correlation coefficients among variables: a) on raw data; b) on weighted data.}
\label{fig:s0v_corr}
\end{figure}

\begin{figure*}[t]
\centering
\subfloat[$\beta$\_Error of $\mathbf{S}$: Mean (green bar) and Variance (black line)]{
  \includegraphics[width=1.6in]{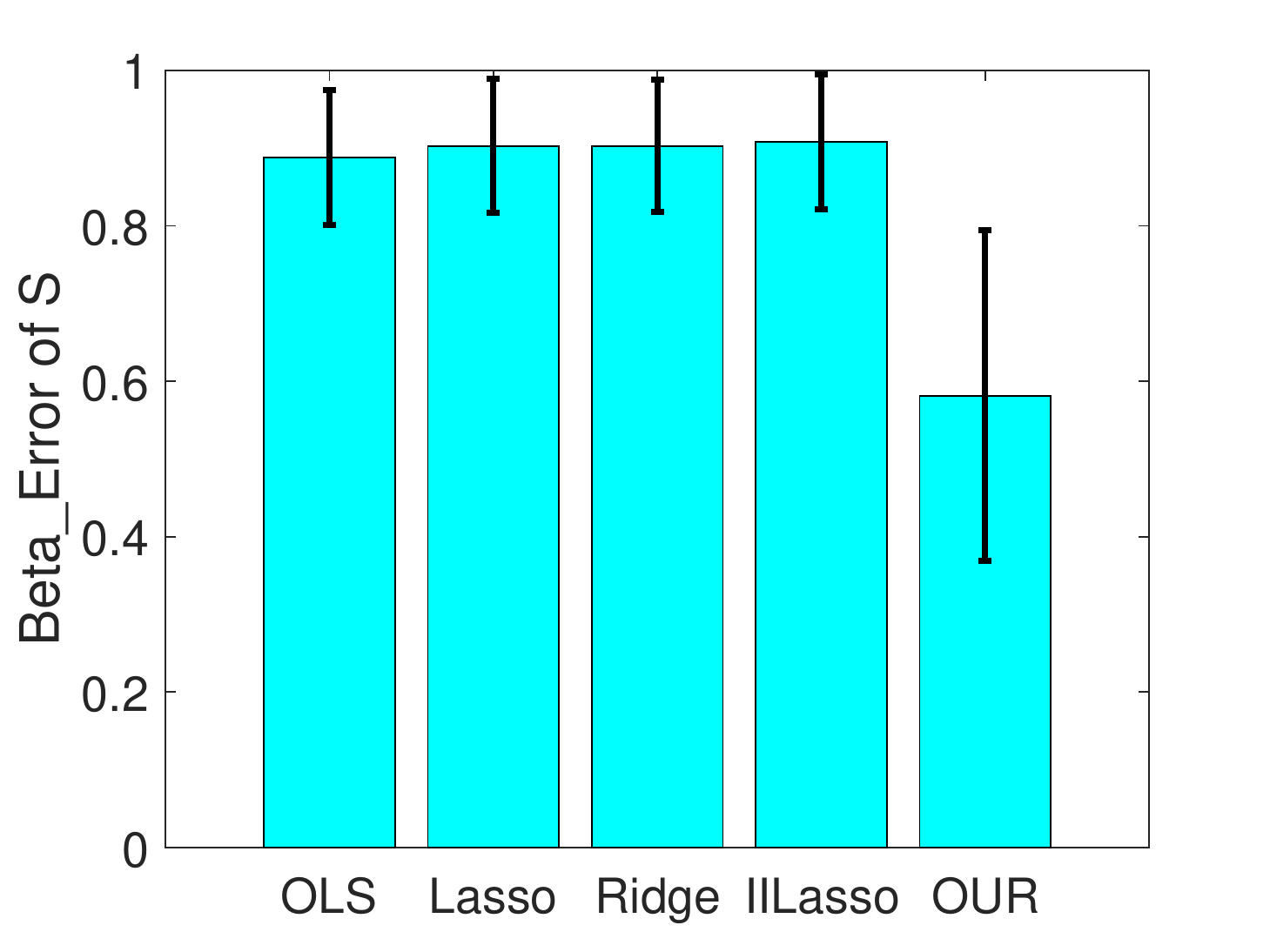}
}
\ \
\subfloat[$\beta$\_Error of $\mathbf{V}$: Mean (green bar) and Variance (black line)]{
  \includegraphics[width=1.6in]{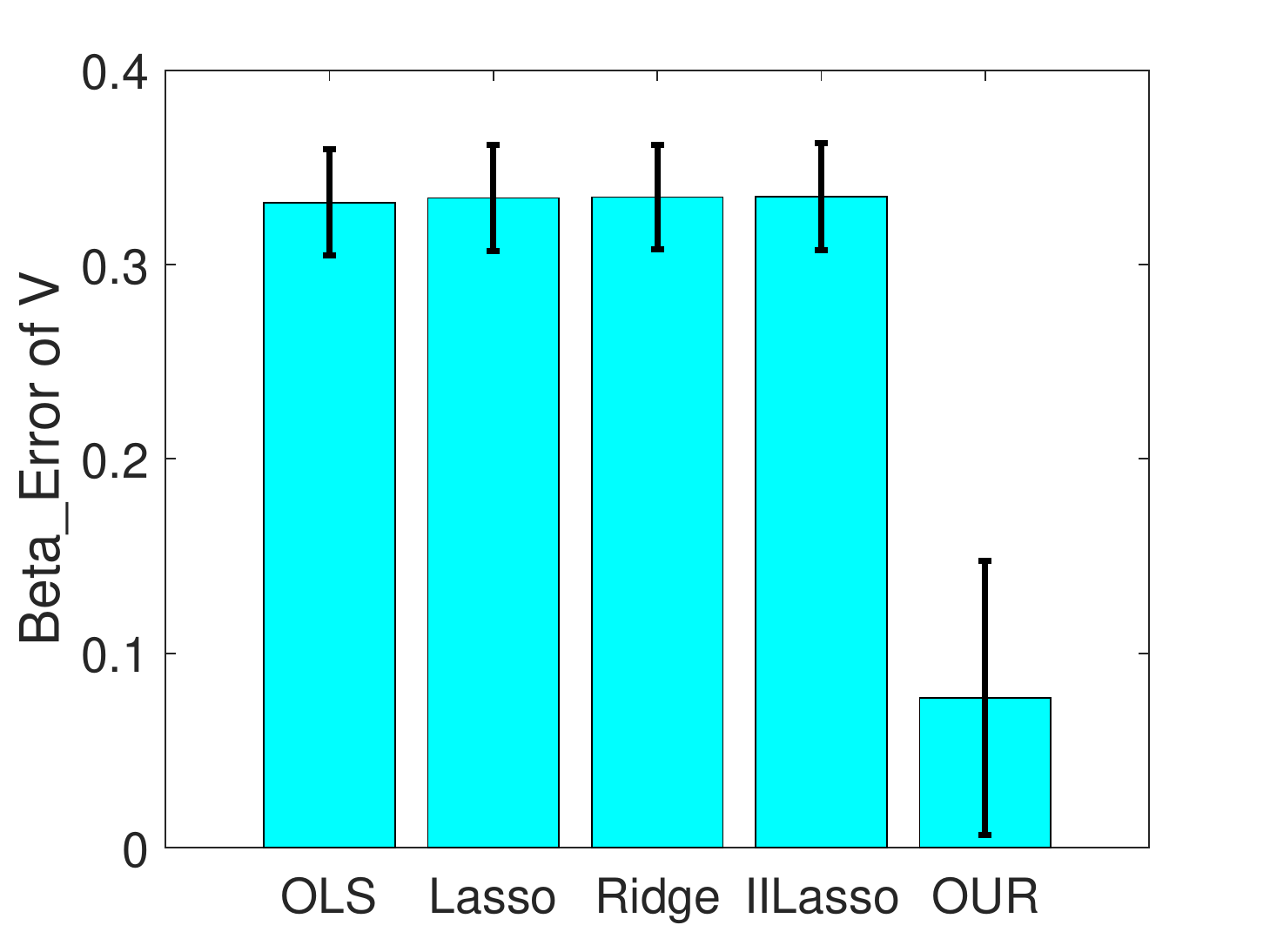}
}
\ \
\subfloat[RMSE over all test environments \label{fig:s0v_rmse}]{
  \includegraphics[width=1.6in]{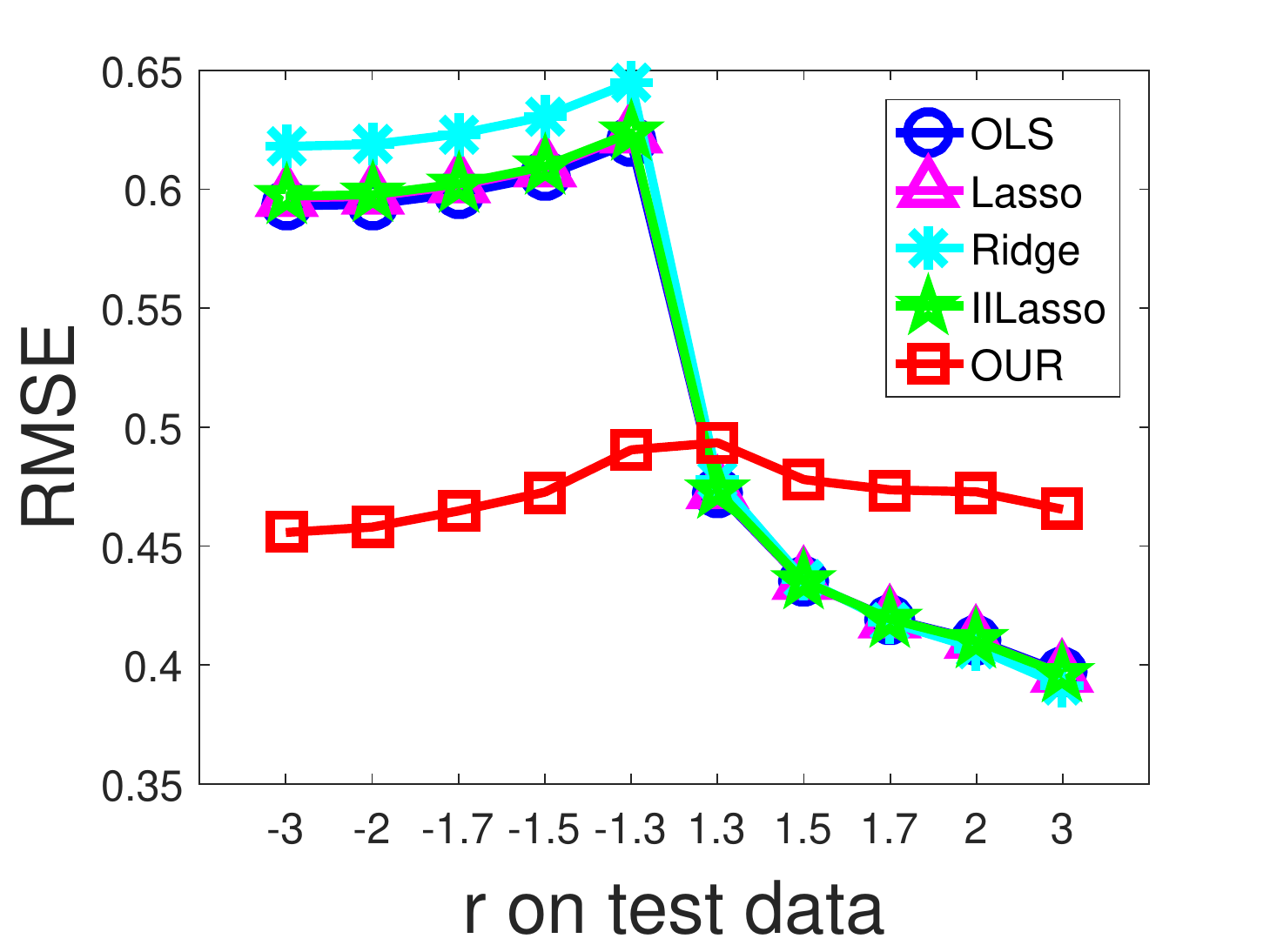}
}
\ \
\subfloat[Average\_Error (green bar) $\&$ Stability Error (black line)]{
  \includegraphics[width=1.6in]{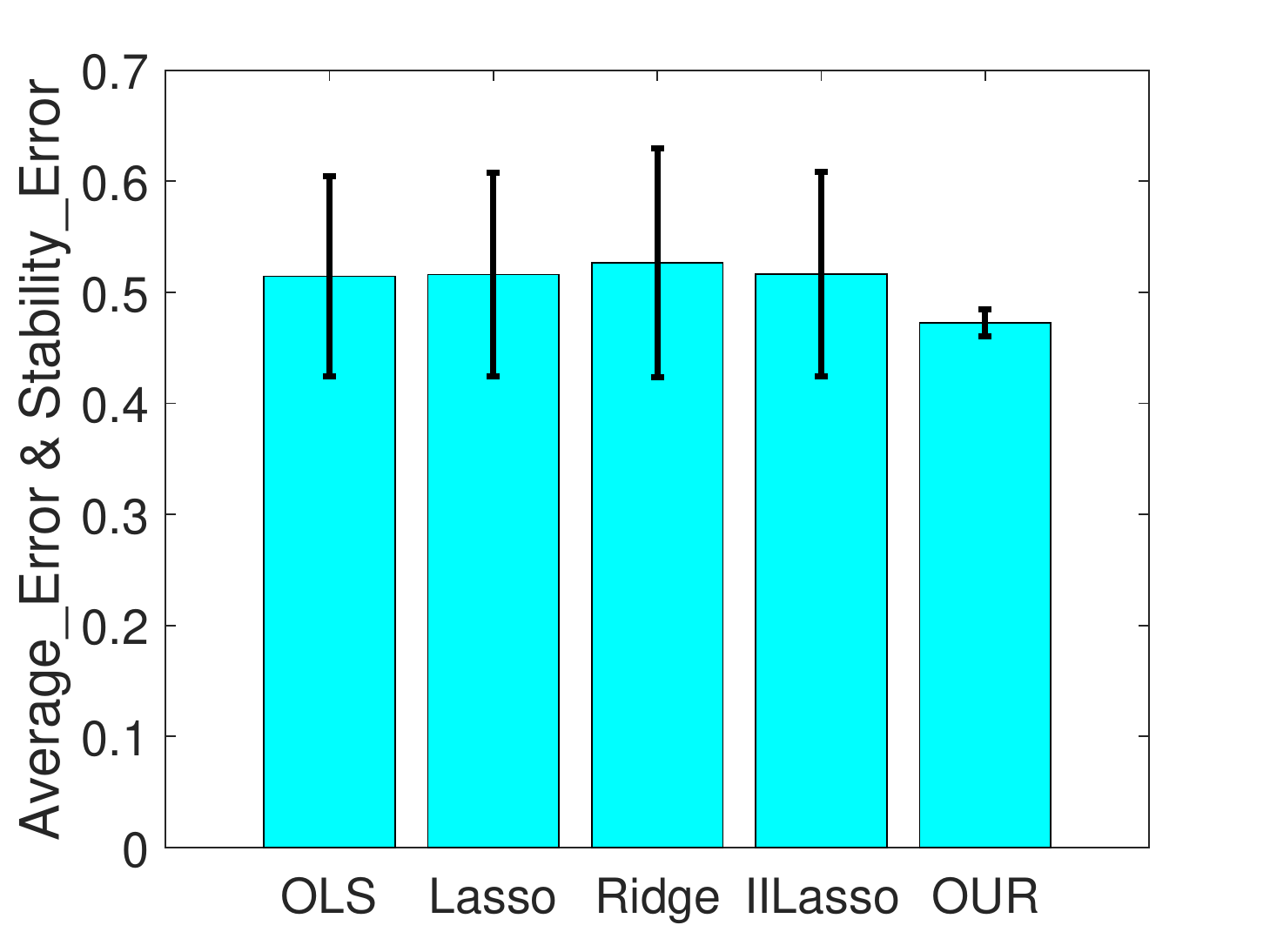}
}
\caption{Results on $\mathbf{S}\perp \mathbf{V}$ with $Y=Y_{poly}$. All the models are trained with $n=2000$, $p=10$ and $r_{train} = 1.7$.}
\label{fig:s0v}
\end{figure*}

\subsubsection{Generating Various Environments}
To test the stability of all algorithms, we need to generate a set of environments, each with a distinct joint distribution $P(\mathbf{X},Y)$, while preserving Assumption \ref{asmp:stable} (and in particular, $P(Y|\mathbf{S})$).
Specifically, we generate different environments in our experiments by varying $P(\mathbf{V}|\mathbf{S})$. For simplification we only change $P(\mathbf{V}_b|\mathbf{S})$ on a subset of unstable features $\mathbf{V}_b \subseteq \mathbf{V}$, where the dimension of $\mathbf{V}_b$ is $0.1*p$.

Specifically, we vary $P(\mathbf{V}_b|\mathbf{S})$ via biased sample selection with a bias rate $r\in[-3,-1)\cup(1,3]$.
For each sample, we select it with probability $Pr = \prod_{\mathbf{V}_{i} \in \mathbf{V}_b}|r|^{-5*D_i}$, where $D_i = |f(\mathbf{S})-sign(r)*\mathbf{V}_{i}|$.
If $r>0$, $sign(r) = 1$; otherwise, $sign(r) = -1$. $f(\mathbf{S})$ is defined in Eq. (\ref{eq:Y_simulation_poly}) or (\ref{eq:Y_simulation_exp}).

Note that $r>1$ corresponds to positive unstable correlation between $Y$ and $\mathbf{V}_b$, while $r<-1$ refers to the negative unstable correlation between $Y$ and $\mathbf{V}_b$.
The higher the value of $|r|$, the stronger correlation between $\mathbf{V}_b$ and $Y$.  Different value of $r$ refers to different environments, hence we can generate different environments by varying $P(\mathbf{V}_b|\mathbf{S})$.

\subsubsection{Experimental Settings}
In experiments, we evaluate the performance of all algorithms from two aspects, including accuracy on parameter estimation and stability on prediction across unknown test data.
To measure the accuracy of parameter estimation, we train all models on one training dataset with a specific bias rate $r_{train}$. We carry out model training for 50 times independently with different training data from the same bias rate $r_{train}$, and report the mean and variance of $\beta\_Error$ on stable features $\mathbf{S}$ and unstable features $\mathbf{V}$.
To evaluate the stability of prediction, we test all models on various test environments with different bias rate $r_{test}\in[-3,-1)\cup(1,3]$.
For each test bias rate $r_{test}$, we generate 50 different test datasets and report the mean of RMSE. With RMSE from all test environments, we report Average$\_$Error and Stability$\_$Error to evaluate the stability of prediction across unknown test environments.

\begin{table*}[t]
% T=logit, Y=linear and nonlinear
\centering
\caption{Experimental results under setting $\mathbf{S} \perp \mathbf{V}$ with $Y=Y_{poly}$ when varying sample size $n$, dimension of variables $p$ and training bias rate $r$. The smaller value of $\beta_S$\_Error, $\beta_V$\_Error, Average\_Error and Stability\_Error, the better.}
\label{tab:results_s0v}
\resizebox{!}{3.5cm}
{
\begin{tabular}{|c|c|c|c|c|c|c|c|c|c|c|c|c|c|c|c|}
  \hline
  \multicolumn{16}{|c|}{\textbf{Scenario 1: varying sample size $n$}}\\
  \hline
$n,p,r$&\multicolumn{5}{|c|}{$n=1000,p=10,r=1.7$}&\multicolumn{5}{|c|}{$n=2000,p=10,r=1.7$}&\multicolumn{5}{|c|}{$n=4000,p=10,r=1.7$}\\
  \hline
  & OLS & Lasso & Ridge & IILasso & Our & OLS & Lasso & Ridge & IILasso & Our & OLS & Lasso & Ridge & IILasso & Our \\
\hline % setting 1
$\beta_{S}$\_Error & 0.892 & 0.907 & 0.907 & 0.912 & \textbf{0.578} & 0.887 & 0.903 & 0.903 & 0.908 & \textbf{0.581} & 0.906 & 0.921 & 0.921 & 0.926 & \textbf{0.614}\\
$\beta_{V}$\_Error & 0.331 & 0.333 & 0.334 & 0.334 & \textbf{0.109} & 0.332 & 0.334 & 0.335 & 0.335 & \textbf{0.077} & 0.338 & 0.340 & 0.341 & 0.341 & \textbf{0.078}\\
Average\_Error      & 0.509 & 0.511 & 0.511 & 0.511 & \textbf{0.476} & 0.514 & 0.516 & 0.527 & 0.516 & \textbf{0.473} & 0.526 & 0.528 & 0.531 & 0.528 & \textbf{0.480}\\
Stability\_Error      & 0.084 & 0.086 & 0.086 & 0.086 & \textbf{0.012} & 0.090 & 0.092 & 0.103 & 0.092 & \textbf{0.012} & 0.104 & 0.105 & 0.108 & 0.106 & \textbf{0.015}\\
\hline
  \hline
\multicolumn{16}{|c|}{\textbf{Scenario 2: varying variables' dimension $p$}}\\
  \hline
$n,p,r$&\multicolumn{5}{|c|}{$n=2000,p=10,r=1.5$}&\multicolumn{5}{|c|}{$n=2000,p=20,r=1.5$}&\multicolumn{5}{|c|}{$n=2000,p=40,r=1.5$}\\
  \hline
  & OLS & Lasso & Ridge & IILasso & Our & OLS & Lasso & Ridge & IILasso & Our & OLS & Lasso & Ridge & IILasso & Our \\
\hline % setting 1
$\beta_{S}$\_Error & 0.618 & 0.628 & 0.630 & 0.632 & \textbf{0.409} & 2.608 & 2.677 & 2.670 & 2.713 & \textbf{1.761} & 8.491 & 8.846 & 8.669 & 8.998 & \textbf{7.800}\\
$\beta_{V}$\_Error & 0.243 & 0.245 & 0.246 & 0.245 & \textbf{0.052} & 0.426 & 0.433 & 0.433 & 0.437 & \textbf{0.260} & 0.661 & 0.684 & 0.673 & 0.694 & \textbf{0.606}\\
Average\_Error      & 0.486 & 0.487 & 0.487 & 0.487 & \textbf{0.476} & 0.523 & 0.527 & 0.539 & 0.529 & \textbf{0.480} & 0.532 & 0.540 & 0.537 & 0.543 & \textbf{0.490}\\
Stability\_Error      & 0.058 & 0.059 & 0.060 & 0.059 & \textbf{0.010} & 0.116 & 0.121 & 0.134 & 0.123 & \textbf{0.014} & 0.138 & 0.148 & 0.145 & 0.153 & \textbf{0.073}\\
\hline
  \hline
\multicolumn{16}{|c|}{\textbf{Scenario 3: varying bias rate $r$ on training data}}\\
  \hline
$n,p,r$&\multicolumn{5}{|c|}{$n=2000,p=10,r=1.5$}&\multicolumn{5}{|c|}{$n=2000,p=10,r=1.7$}&\multicolumn{5}{|c|}{$n=2000,p=10,r=2.0$}\\
  \hline
  & OLS & Lasso & Ridge & IILasso & Our & OLS & Lasso & Ridge & IILasso & Our & OLS & Lasso & Ridge & IILasso & Our \\
\hline % setting 1
$\beta_{S}$\_Error & 0.618 & 0.628 & 0.630 & 0.632 & \textbf{0.409} & 0.887 & 0.903 & 0.903 & 0.908 & \textbf{0.581} & 1.232 & 1.249 & 1.245 & 1.257 & \textbf{0.651}\\
$\beta_{V}$\_Error & 0.243 & 0.245 & 0.246 & 0.245 & \textbf{0.052} & 0.332 & 0.334 & 0.335 & 0.335 & \textbf{0.077} & 0.441 & 0.444 & 0.443 & 0.445 & \textbf{0.119}\\
Average\_Error      & 0.486 & 0.487 & 0.487 & 0.487 & \textbf{0.476} & 0.514 & 0.516 & 0.527 & 0.516 & \textbf{0.473} & 0.568 & 0.571 & 0.571 & 0.571 & \textbf{0.476}\\
Stability\_Error      & 0.058 & 0.059 & 0.060 & 0.059 & \textbf{0.010} & 0.090 & 0.092 & 0.103 & 0.092 & \textbf{0.012} & 0.144 & 0.147 & 0.147 & 0.147 & \textbf{0.008}\\
\hline
\end{tabular}
}
\end{table*}

\subsubsection{Results}
Before reporting the experimental results, we demonstrate the Pearson correlation coefficients between any two variables on both raw data and the weighed data by our algorithm in Figure \ref{fig:s0v_corr}.
From the figures, we can find that in the raw data, the unstable features $\mathbf{V}_5$ is correlated with some stable features $\mathbf{S}$, and highly correlated with both omitted nonlinear term $g$ and outcome $Y$.
Hence, the estimated coefficient of $\mathbf{V}_5$ in the baselines would be large, which should be $zero$ in a correctly specified model, leading to unstable prediction.
In the weighted data, the sample weights learnt from our algorithm can clearly remove the correlation among predictors $\mathbf{X}$.
Moreover, the unstable correlation between $\mathbf{V}_5$ and $g$ are significantly reduced, which is helpful to reduce the unstable correlation between $\mathbf{V}_5$ and $Y$, and then the correlations between stable features $\mathbf{S}$ and $Y$ conditional on $\mathbf{V}$ are enhanced.
Hence, our algorithm can estimate the coefficient of both $\mathbf{S}$ and $\mathbf{V}$ more precisely.
This is the key reason that our algorithm can make more stable predictions across unknown test environments.

We report the results of parameter estimation and stable prediction under setting $\mathbf{S}\perp \mathbf{V}$ with $Y=Y_{poly}$ in Figure \ref{fig:s0v} and Table \ref{tab:results_s0v}. To save space, the experimental results of settings $\mathbf{S}\rightarrow \mathbf{V}$ and $\mathbf{V}\rightarrow \mathbf{S}$ with $Y=Y_{poly}$, and results with $Y=Y_{exp}$ are reported in online Appendix.
From the results, we have following observations and analysis:
\begin{itemize}%[leftmargin=0.7cm]
\item \par \noindent OLS cannot address the stable prediction problem. The reason is that OLS is biased on both $\beta_S$ and $\beta_V$ estimation as we discussed in the theoretical section. Moreover, OLS will often predict large effects of the unstable features, which leads to instability across environments.
\item \par \noindent Lasso, Ridge and IILasso perform even worse than OLS, since their regularizers will generally estimate larger coefficients on the unstable features $\mathbf{V}_b$. For example, Lasso selects a only a subset of predictors and exacerbates the omitted variables problem that already exists in our basic setup.
\item \par \noindent Comparing with baselines, our algorithm achieves more stable prediction across different settings. By reducing the correlation among all predictors, our algorithm avoids using unstable features to proxy for omitted nonlinear functions of the stable features, ensuring less bias in the estimation of the effect of both stable features and unstable features. Hence, improve the stability of prediction.
\item \par \noindent The performance of our algorithm is worse than baseline when $r_{test}>1.3$ on test data in Fig. \ref{fig:s0v_rmse}, but much better than baselines when $r_{test}<-1.3$. This is because the correlations between $\mathbf{V}_b$ and $Y$ are similar between training data ($r_{train}=1.7$) and test data when $r_{test}>1.3$, and that correlation can be exploited in prediction; in this setting, $\mathbf{V}$ is useful to proxy for omitted functions of $\mathbf{S}$. However, when $r_{test}<-1.3$, using $\mathbf{V}$ for prediction creates too much instability.
\item \par \noindent By varying the sample size $n$, dimension of variables $p$, training bias rate $r_{train}$ and the form of missing nonlinear and interaction terms, our algorithm is consistently outperform than baselines on parameter estimation and stable prediction across unknown test data.
\end{itemize}
Overall, our proposed DWR algorithm can be applied to address the problem of stable prediction with model misspecification and agnostic distribution shift.

\subsubsection{Parameter Analysis}

\begin{figure}[t]
\centering
\subfloat[$Average\_Error$ v.s. $\lambda_2$]{
  \includegraphics[width=1.6in]{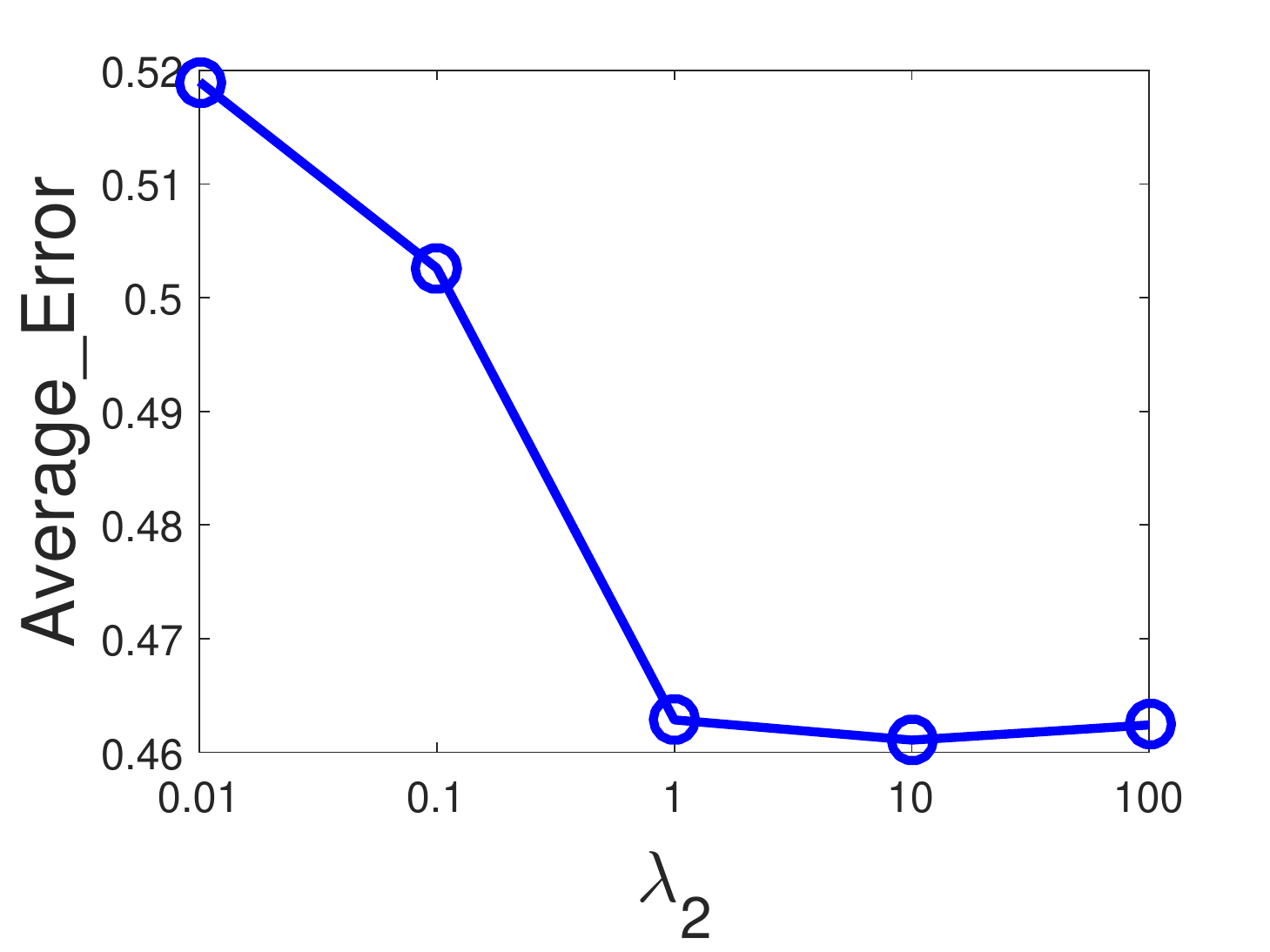}
}
\subfloat[$Stability\_Error$ v.s. $\lambda_2$]{
  \includegraphics[width=1.6in]{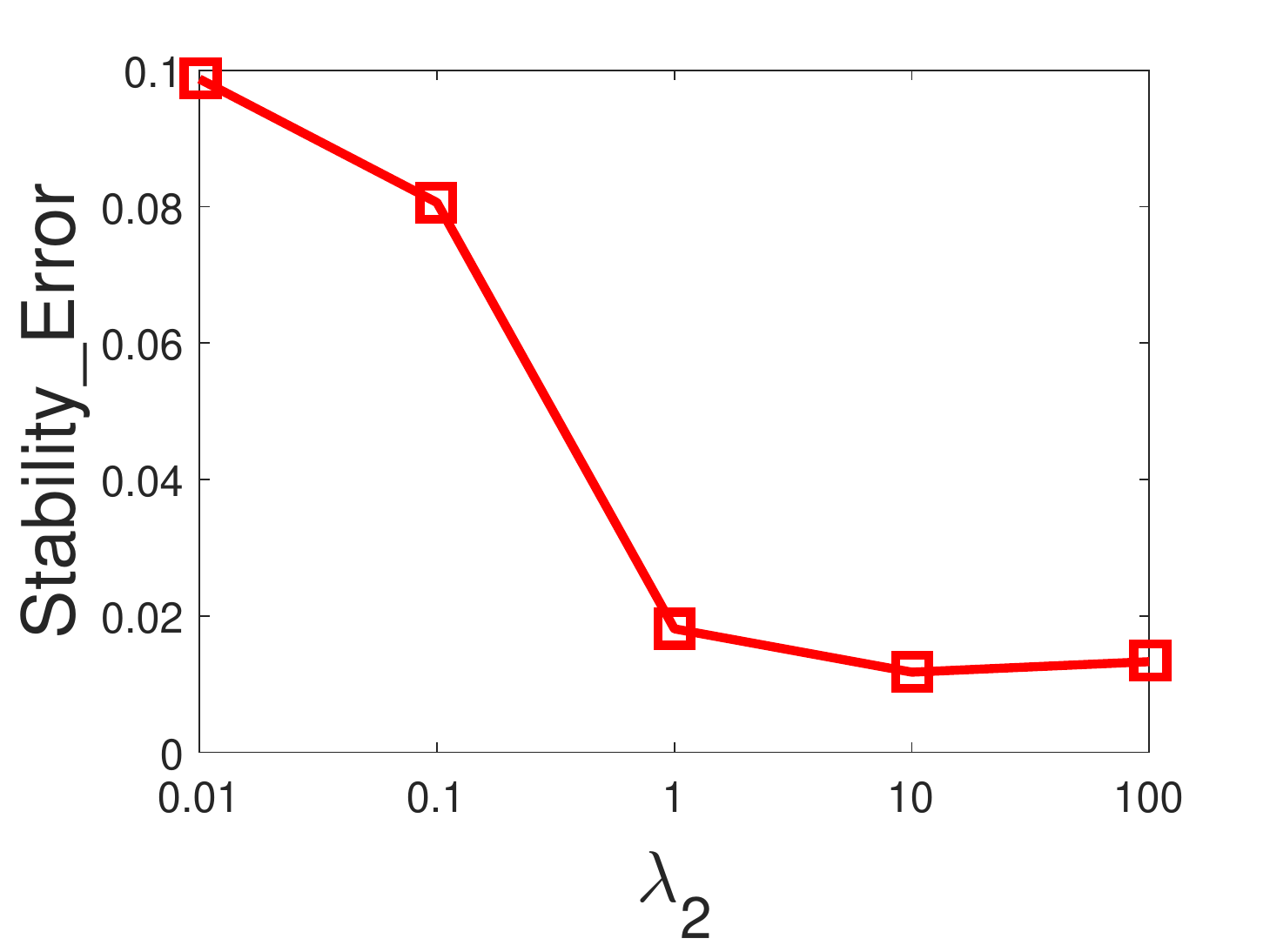}
}
\caption{The effect of hype-parameter $\lambda_2$.}
\label{fig:hype_parameter}
\end{figure}

In our DWR algorithm, we have some hype-parameters, including $\lambda_1$ for constraining the sparsity of regression coefficient, $\lambda_2$ for constraining the error of decorrelation regularizer, $\lambda_3$ for constraining the variance of the sample weights, and $\lambda_4$ for constraining the sum of sample weights to $n$.
In our experiments, we tuned these parameters with cross validation by grid searching, and each parameter is uniformly varied from $\{0.01,0.1,1,10,100\}$.
In Figure \ref{fig:hype_parameter}, we displayed $Average\_Error$ and $Stability\_Error$ with respect to $\lambda_2$. From the figures, we can find that when $\lambda_2<10$,  $Average\_Error$ and $Stability\_Error$ monotonically decrease as we increase the value of hype-parameter $\lambda_2$. But when $\lambda_2 >10$, those errors will slightly increase as we keep increasing the value of hype-parameter $\lambda_2$.

\subsection{Experiments on Real World Data}

\subsubsection{Datasets and Experimental Setting}

We collected air pollutant data and meteorological data from the U.S. EPA's Air Quality System (AQS) database,\footnote{\url{https://www.epa.gov/outdoor-air-quality-data}} which has been widely used for model evaluation \cite{yahya2017decadal,zhu2018machine}.
The air pollutant data in this study is PM$_{10}$, and the meteorological variables are those would affect the air pollutant concentrations, including air temperature, relative humidity, pressure, wind speed and direction.

In our experiments, we let the outcome variable $Y$ be pollution PM$_{10}$, and set the meteorological features as the observed variables $\mathbf{X}$.
To test the stability of all algorithms, we collected data from 10 different states in the U.S., where the states correspond to the different environments from the theory.
Considering a practical setting where a researcher has a single data set and wishes to train a model that can then be applied to other related settings, in our experiments, we trained all models with data from State 1, validated with data from States 1 to 4, finally tested them on all 10 States.

To demonstrate the distribution difference between any two environments $e=i$ and $e=j$, we adopt the distribution distance\footnote{Variable's distribution can be uniquely determined by all the collections of its moments. Here, we only consider the first moment. Other metrics can also be applied to measure distribution distance, for example, KL-divergence. We leave it in the future work.} between observed variables $\mathbf{X}$ as a metric with following definition:
\begin{eqnarray}
\nonumber \Scale[1.0]{Distribution\_Distance(i,j) = \sum_{k = 1}^{p}\|\overline{\mathbf{X}}_{e=i}-\overline{\mathbf{X}}_{e=j}\|,}
\end{eqnarray}
where $p$ refers to the dimension of variables, and $\overline{\mathbf{X}}_{e=i}$ represents the mean value of variables $\mathbf{X}$ in environment $i$.
\subsubsection{Results}

\begin{figure}[t]
\centering
\subfloat[RMSE v.s. Distribution Distance \label{fig:realdata_rmse}]{
  \includegraphics[width=1.7in]{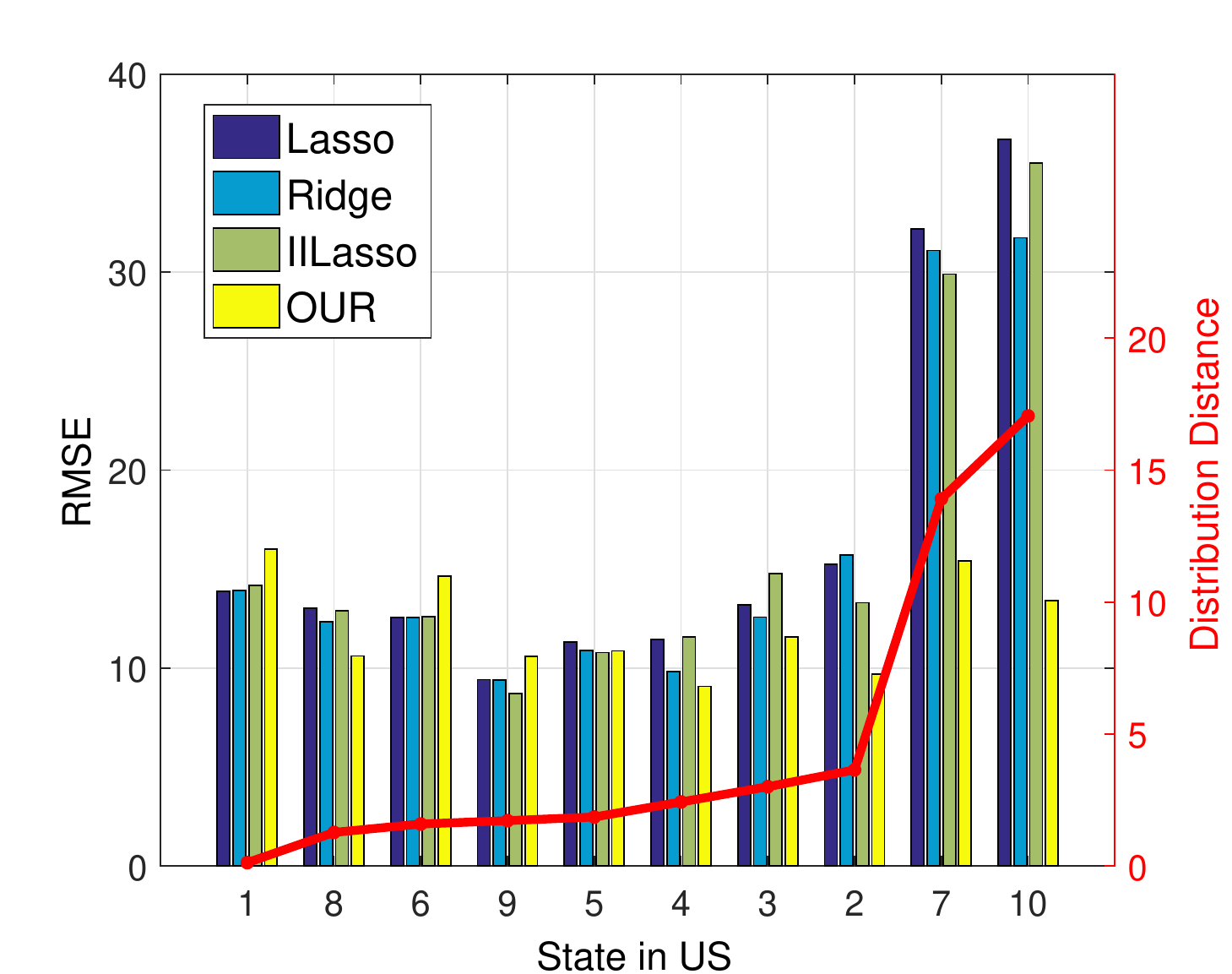}
}
\subfloat[Average$\_$Error (green bar) $\&$ Stability$\_$Error (black line) \label{fig:realdata_stability}]{
  \includegraphics[width=1.7in]{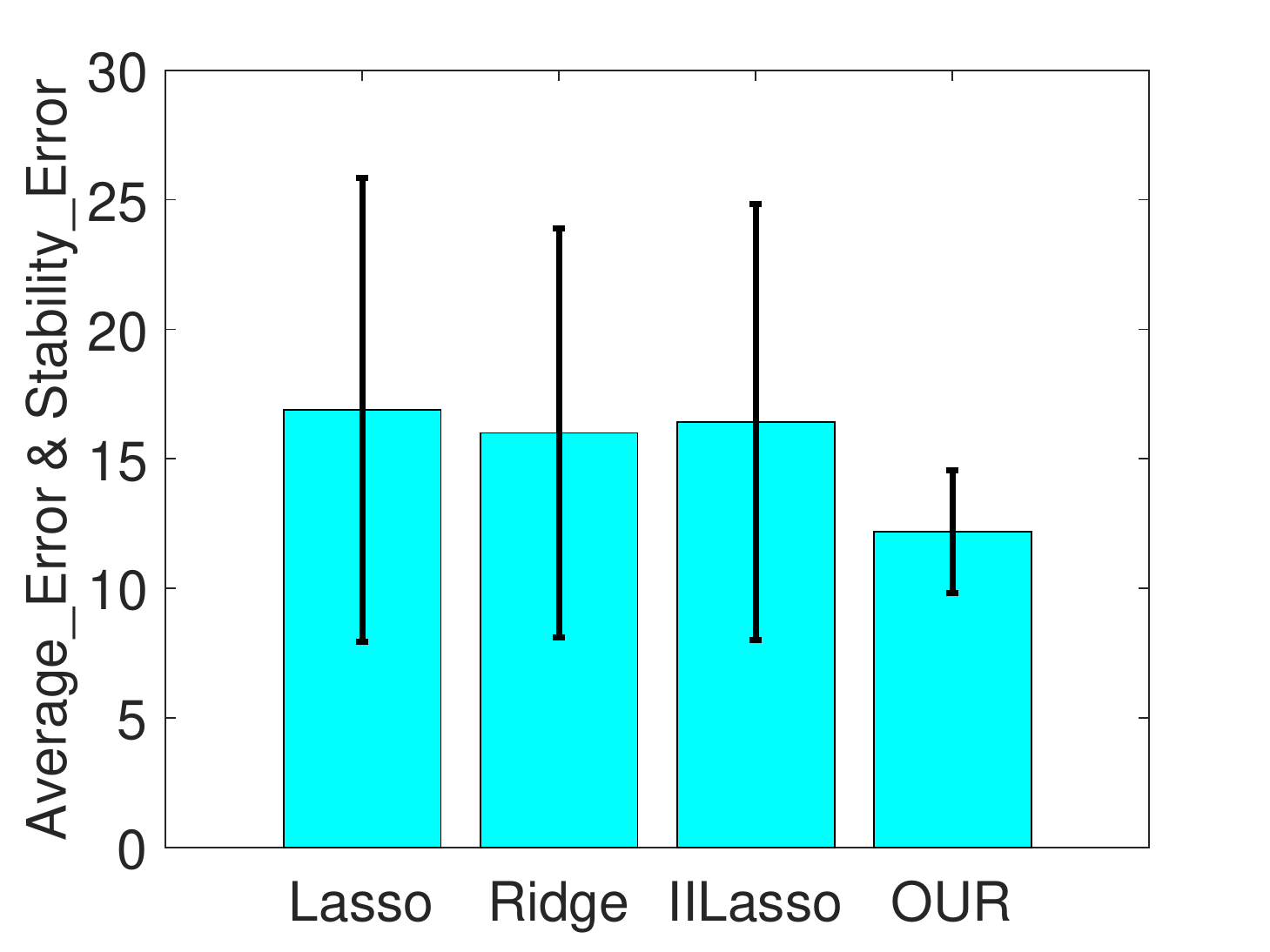}
}
\caption{Air quality prediction across different States in US. Models are trained on State 1. The red line represents the distance between training and test distribution.}
\label{fig:realdata}
\end{figure}

\hide{
\begin{figure}[t]
\centering
\subfloat{
  \includegraphics[width=2.6in]{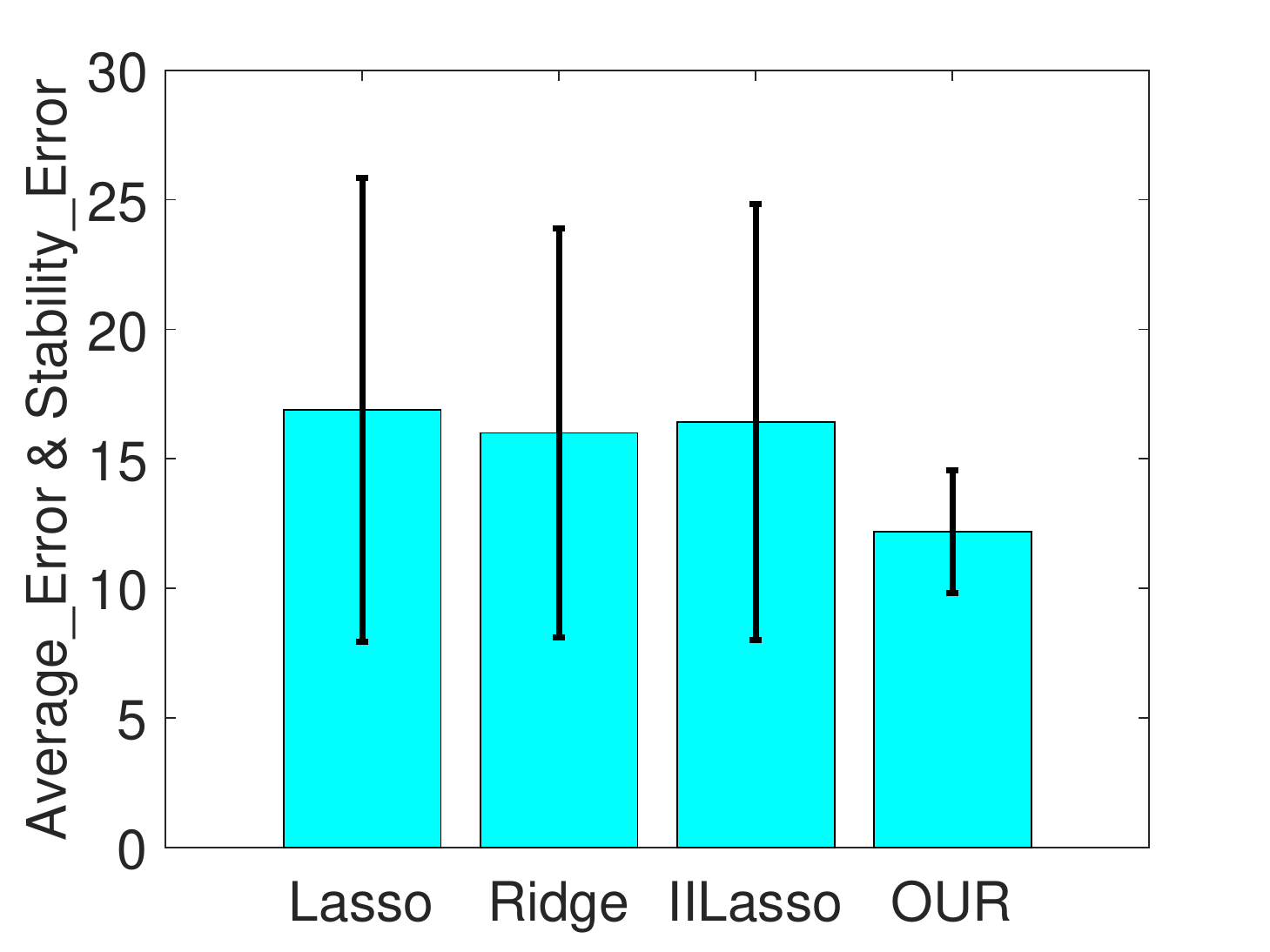}
}
\caption{Average$\_$Error and Stability$\_$Error of air quality prediction across different States in US.}
\label{fig:realdata_stability}
\end{figure}
}

We report the results of RMSE on air quality prediction over all 10 States in Fig. \ref{fig:realdata_rmse}, where we merged OLS method into Lasso by allowing its hype-parameter to be $zero$ during model training.
The results show that the performance of our algorithm is worse than baselines when the distribution distance between training and test environments is small; in that case, we introduce variance by reweighting the data away from the distribution that approximates both training and test sets. But our algorithm's performance improves relative to the baseline and ultimately becomes better than baseline as the distribution distance increases.

To explicitly demonstrate the advantage of our proposed algorithm, we report Average\_Error and Stability\_Error in Fig. \ref{fig:realdata_stability}. The results show that our algorithm makes the most stable prediction with agnostic distribution shift on test data.

\section{Conclusion}

In this paper, we focus on how to facilitate a stable prediction across unknown test data, where we are concerned about two problems that together lead to instability: model misspecification, and agnostic distribution shift between training and test data.
We proved that our algorithm can improve the accuracy of parameter estimation and stability on prediction from both theoretical analysis and empirical experiments. The experimental results on both synthetic and real-world datasets demonstrate that our algorithm outperforms the baselines for stable prediction across unknown test environments, when the correlation among covariates varies substantially across those environments.

\section{Acknowledgement}

%This work was supported by National Key Research and Development Program of China (2018AAA0101900).
This work was supported by National Key Research and Development Program of China (No. 2018AAA0102004, No. 2018AAA0101900), National Natural Science Foundation of China (No. 61772304, No. 61521002, No. 61531006, No. U1611461), Beijing Academy of Artificial Intelligence (BAAI).
Susan Athey's research was supported by Sloan foundation and Office of Naval Research grant N00014-17-1-2131.
Bo Li's research was supported by the Tsinghua University Initiative Scientific Research Grant, No. 20165080091; National Natural Science Foundation of China, No. 71490723 and No. 71432004; Science Foundation of Ministry of Education of China, No. 16JJD630006.
All authors of this paper are corresponding authors.
All opinions, findings, and conclusions in this paper are those of the authors and do not necessarily reflect
the views of the funding agencies.

%\newpage
{
%\small
\bibliographystyle{aaai}
\bibliography{aaai20}  % sigproc.bib is the name of the Bibliography in this case
}

\noindent \textbf{Appendix}: The online appendix and supplementary materials are available at http://kunkuang.github.io or https://www.dropbox.com/s/1q0brkc2bnehhfo/paper-aaai20-Supplementary.

\end{document}